\documentclass{article} % For LaTeX2e
\usepackage{iclr2026_conference,times}

\usepackage{amsmath,amsfonts,bm}

\def\eqref#1{equation~\ref{#1}}
\def\1{\bm{1}}

\DeclareMathAlphabet{\mathsfit}{\encodingdefault}{\sfdefault}{m}{sl}
\SetMathAlphabet{\mathsfit}{bold}{\encodingdefault}{\sfdefault}{bx}{n}

\usepackage{hyperref}
\usepackage{url}
\usepackage{amssymb}
\usepackage{booktabs}
\usepackage{graphicx}
\usepackage{multirow}
\usepackage{xcolor,colortbl}
\usepackage{wrapfig}
\usepackage[breakable]{tcolorbox}

\definecolor{tabhighlight}{HTML}{E3F3FC}
\definecolor{arxivabstractbg}{HTML}{E3F3FC}
\definecolor{arxivaccent}{HTML}{005BAC}
\colorlet{arxivabstractborder}{arxivaccent}

\newcommand{\ours}{\text{Viscot-steer}}

\newtcolorbox{promptbox}[1]{
    breakable,
    colback=black!1,
    colframe=black!25,
    colbacktitle=tabhighlight,
    coltitle=black,
    fonttitle=\small\bfseries,
    fontupper=\small\ttfamily\raggedright,
    boxrule=0.4pt,
    arc=2pt,
    left=8pt,
    right=8pt,
    top=6pt,
    bottom=6pt,
    before skip=10pt,
    after skip=10pt,
    title={#1}
}

\usepackage{hyperref}
\hypersetup{
    colorlinks=true,
    linkcolor=red,
    filecolor=magenta,
    urlcolor=cyan,
    citecolor=arxivaccent,
}
\RequirePackage[shortlabels,inline]{enumitem}
\setlist[itemize]{noitemsep,leftmargin=1.5em,topsep=0em}
\setlist[enumerate]{noitemsep,leftmargin=1em,topsep=0em}

\title{\ours: Disentangling and Transferring Visual Chain-of-Thought via Activation Steering}

\author{%
Ningning Wang\textsuperscript{1},
Beier Zhu\textsuperscript{1,\correspondingauthor},
Xingyu Zhu\textsuperscript{2},
Kaiyun Yang\textsuperscript{1},
Xun Yang\textsuperscript{1,\correspondingauthor}\\[0.6em]
{\small\textsuperscript{1}University of Science and Technology of China}\\[0.2em]
{\small\textsuperscript{2}National University of Singapore}\\[0.3em]
{\footnotesize\correspondingauthor\enspace Corresponding authors}
}

\iclrfinalcopy
\input{arxiv-style}
\begin{document}

\maketitle
% Do not imply conference acceptance on the arXiv preprint.
\fancyhead{}
% \begin{abstract}
% Visual chain-of-thought (visual-CoT) enables large vision-language models (LVLMs) to ground intermediate reasoning in task-relevant visual evidence, thereby improving their ability to solve complex visual reasoning tasks.
% However, training a visual-CoT model relies on carefully curated reasoning and grounding data together with supervised fine-tuning, reinforcement learning, or both, and this costly process must be repeated whenever the underlying base model is upgraded or replaced.
% In this paper, we propose \ours, a training-free activation steering framework for transferring visual-CoT capabilities across heterogeneous LVLMs.
% Visual-CoT comprises two complementary capabilities, grounding and reasoning, that operate at different model depths.
% Accordingly, \ours~disentangles visual-CoT into grounding and reasoning capability vectors, extracts them from capability-specific layers of a source visual-CoT model, aligns them with the target activation space, and injects each vector into the corresponding target layer only during the generation phase in which that capability is needed.
% Experiments on visual reasoning tasks spanning real-world and mathematical scenarios demonstrate effective visual-CoT capability transfer from different visual-CoT models to LVLMs with diverse architectures, achieving stronger overall performance than representative capability-transfer baselines.
% \end{abstract}
\begin{abstract}
Visual chain-of-thought (visual-CoT), also known as ``thinking with images'', equips LVLMs with the ability to locate and use key visual evidence for reasoning.
However, acquiring such visual-CoT capabilities requires costly post-training and tightly couples them to a specific backbone, making transfer to other LVLMs difficult.
Existing cross-model transfer methods typically treat visual-CoT as a monolithic capability, using a unified representation that ignores the
functional distinction between grounding and reasoning and introduces stage-irrelevant interventions.
To address this problem, we propose \ours, a training-free activation steering framework for transferring visual-CoT capabilities across heterogeneous LVLMs.
Specifically, \ours~first disentangles visual-CoT into grounding and reasoning
vectors and extracts them from the source-model layers associated with the
corresponding capabilities. It then aligns the vectors with the activation
space of the target model and steers the grounding and reasoning layers during
bounding-box prediction and reasoning-trace generation, respectively.
Across benchmarks covering high-resolution visual understanding, general visual perception and reasoning, and mathematical visual reasoning, \ours~transfers capabilities from different visual-CoT models to architecturally diverse LVLMs and outperforms the strongest capability-transfer baseline by an average of 1.86 percentage points on the primary evaluation metrics.
\end{abstract}

\section{Introduction}
\label{sec:intro}
Visual chain-of-thought (visual-CoT), also known as “thinking with images,” enhances multimodal reasoning by interleaving textual reasoning with explicit grounding in task-relevant visual evidence. Existing methods acquire these capabilities by constructing multimodal reasoning trajectories with grounding supervision and post-training a specific base model through SFT, RL, or both~\citep{zheng2026deepeyes,pixelreasoner,fan2025grit}. This tightly couples visual-CoT capabilities to the trained backbone: moving to a different base LVLM requires repeating the post-training process, incurring substantial training cost.
This raises a natural question: \textit{can visual-CoT capabilities learned by one model be transferred directly to a new base LVLM, without repeating costly post-training?}

Activation steering offers a promising route toward this goal. Its key premise is that a model capability is reflected in systematic changes in its internal activations. By comparing hidden states elicited in a source model with and without the capability of interest under matched inputs, one can isolate its associated activation direction. Cross-model transfer then can be achieved by aligning this direction with the target model’s activation space and injecting it during inference, without updating model parameters. Recent work has demonstrated this paradigm for cross-model transfer across tasks, modalities, and model architectures~\citep{wang2026commandv,gan-etal-2026-textual,zhan2025l2v}.

\begin{figure*}[t]
    \centering
\includegraphics[width=0.95\textwidth]{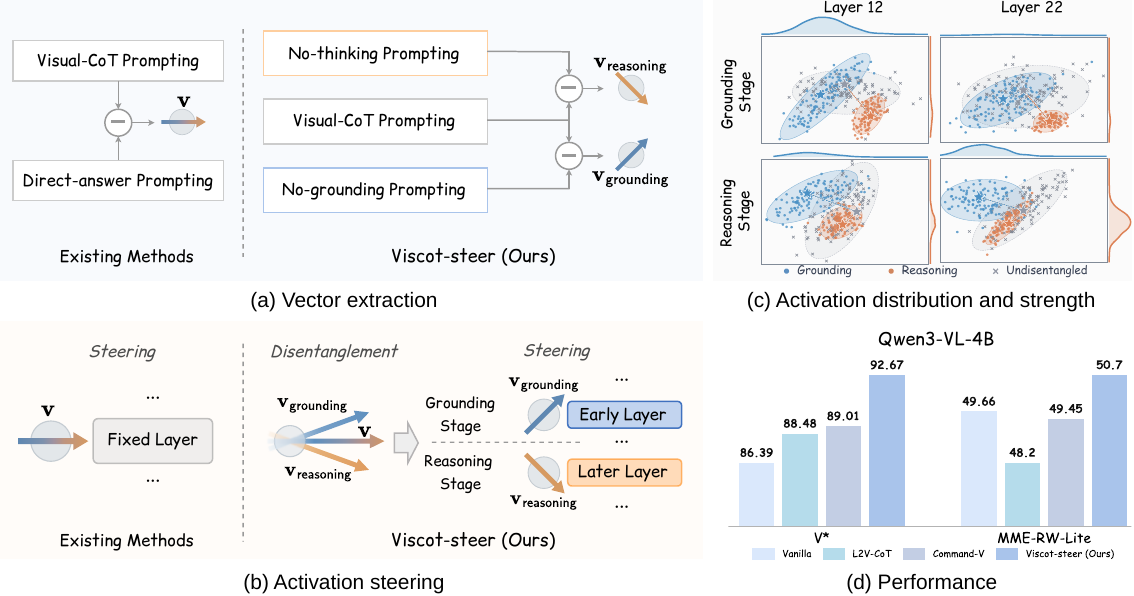}
    \caption{\textbf{Disentangling and transferring visual-CoT with our \ours}.
(a) Existing methods extract a unified visual-CoT vector, whereas our \ours~extracts separate grounding and reasoning vectors using capability-specific contrastive prompts.
(b) Existing methods inject a unified vector at a fixed layer, whereas our \ours~steers each vector at its corresponding layer and generation phase.
(c) Activation distributions and strengths of grounding, reasoning, and undisentangled visual-CoT at the early (12th) and later (22nd) layers during the grounding and reasoning stages.
(d) Our \ours~outperforms the existing baselines by a substantial margin.}
    \label{fig:teaser}
\end{figure*}

Despite this progress, existing cross-model steering methods typically represent
a transferable capability with a single unified direction extracted from a \textbf{fixed} layer and apply it \textbf{uniformly} throughout generation. This treatment is poorly
suited to visual-CoT, which couples functionally \textbf{distinct grounding and reasoning capabilities}. As shown in Fig.~\ref{fig:teaser}(c), the undisentangled visual-CoT direction shifts toward grounding during the grounding stage and toward reasoning during the reasoning stage; moreover, grounding is more active at layer 12,
whereas reasoning is more active at layer 22. These stage- and layer-dependent patterns motivate disentangling visual-CoT and steering each capability only at its specialized layer and corresponding generation phase.

To address this problem, we propose \ours, a training-free activation steering framework for transferring visual-CoT capabilities across heterogeneous LVLMs.
\ours~consists of two stages. 
\textbf{(1) Disentangled capability extraction (what to transfer).}
We construct capability-specific contrastive prompt pairs and extract their
activation differences at specialized source layers, yielding separate grounding
and reasoning capability vectors (Figure~\ref{fig:teaser}(a)).
\textbf{(2) Phase-aware capability steering (where \& when to steer).}
After aligning the vectors to the target activation space, we inject the reasoning vector into reasoning-specific layers during reasoning-trace generation and the grounding vector into grounding-specific layers during bounding-box prediction
(Figure~\ref{fig:teaser}(b)). The answer phase remains unsteered, and no model parameters are updated during
the transfer process.

We compare \ours~with three capability-transfer baselines, including L2V-CoT~\citep{zhan2025l2v}, on benchmarks spanning \textit{high-resolution visual understanding, general visual perception and reasoning, and mathematical visual reasoning}.
Across 25 comparable model--metric combinations, \ours~achieves the best result in 21 combinations and remains close to the strongest baseline in the other four, improving over the strongest capability-transfer baseline by 1.86 percentage points on average.
Ablation studies further demonstrate the effectiveness of capability disentanglement and phase-aware routing.
Experiments with different source visual-CoT models provide further validation.
Our main contributions are:
\begin{itemize}
\item \textbf{(Analysis)} Through layer- and stage-wise analyses, we show that grounding and reasoning specialize at different layers and serve bounding-box prediction and reasoning-trace generation, respectively; unified steering therefore causes cross-stage interference.
\item \textbf{(Method)} We propose \ours, which separately extracts and aligns grounding and reasoning vectors and steers the corresponding capabilities at the required model layers and generation stages, enabling training-free visual-CoT transfer across heterogeneous LVLMs.
\item \textbf{(Performance)} Across five LVLMs and five visual reasoning benchmarks, \ours~outperforms the strongest capability-transfer baseline by an average of 1.86 pp over 25 comparable settings, with further validation across different source visual-CoT models. 
\end{itemize}

\section{Related Work}
\label{sec:related}

\noindent\textbf{Visual-CoT (thinking with images).}
Visual chain-of-thought (visual-CoT) enhances the reasoning capabilities of LVLMs by interleaving language-based reasoning with explicit localization of relevant visual evidence~\citep{shao2024visual,lai2025mini-o3,zhang2026thyme,hu2024visualsketchpad}.
For example, DeepEyes~\citep{zheng2026deepeyes}, UV-CoT~\citep{zhao2025uvcot}, NV-CoT~\citep{zhao2026nvcot} and Pixel-Reasoner~\citep{pixelreasoner} feed crops of localized image regions back to the LVLM for further reasoning, while Visual grounded reasoning~\citep{wang2026vgr} and GRIT~\citep{fan2025grit} associate intermediate reasoning steps with explicit points or bounding-boxes.
Training visual-CoT typically requires carefully curated data and optimization through SFT, RL, or both, and must be repeated when the backbone changes. In contrast, our work transfers visual-CoT capabilities from a trained source model to architecturally heterogeneous LVLMs without additional training. 

\noindent\textbf{Activation steering and cross-model transfer.}
Activation steering controls model behavior through inference-time interventions in intermediate activations and has been widely applied to safety, factuality, generation style, and reasoning~\citep{turner2023steering,li2024inferenceintervent,rimsky-etal-2024-steering,arditi2024refusal,tang-etal-2025-unlocking,zhao2025activation,wu-etal-2025-sharp,wang-etal-2024-inferaligner}. While early approaches primarily steer capabilities within the same model, recent studies extend activation steering to cross-model capability transfer across tasks, modalities, and model architectures~\citep{wang2026commandv,zhan2025l2v,tang-etal-2025-unlocking,tang2025enhancingcrosstasktransferlarge,huang-etal-2025-cross}. However, these methods typically represent a transferred capability with a unified direction, which is inadequate for visual-CoT because grounding and reasoning exhibit distinct activation directions, layer-wise specialization, and generation-time roles. We instead disentangle visual-CoT into grounding and reasoning capability vectors and steer each at its corresponding model depth and generation stage.

\section{Method}
\begin{figure*}[t]
    \centering
\includegraphics[width=0.98\textwidth]{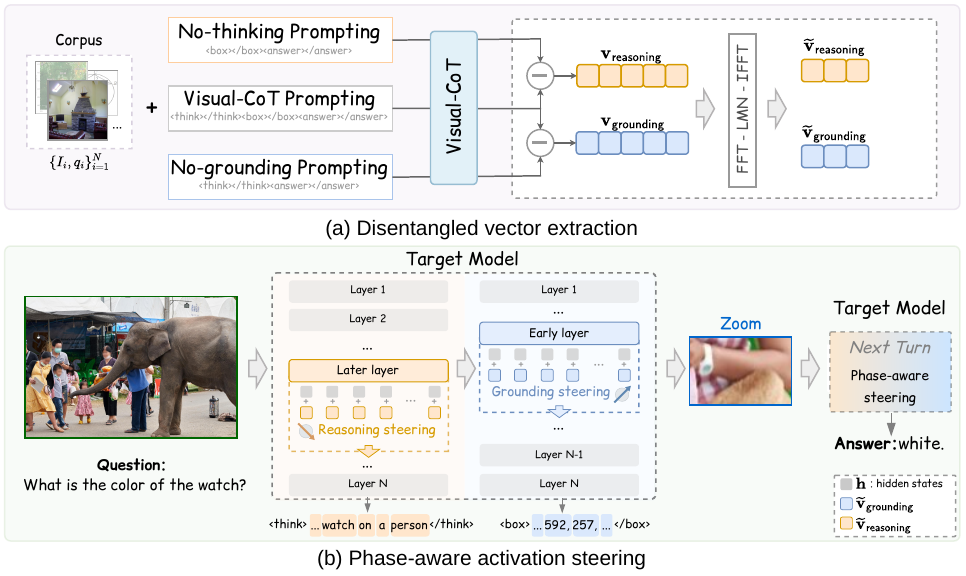}
    \caption{\textbf{Overview of our \ours~framework.} (a) Disentangled vector extraction. Disentangled reasoning and grounding vectors are extracted from capability-specific contrastive prompts. (b) Phase-aware activation steering. The reasoning and grounding vectors are routed only to their corresponding generation phases and injected at capability-specific target layers.} 
    \label{fig:overview}
\end{figure*}

We first formalize visual-CoT reasoning (\S\ref{sec:preliminaries}).
Our pipeline first extracts disentangled grounding and reasoning capability vectors from a source visual-CoT LVLM (\S\ref{sec:capability_extraction}).
It then aligns the extracted vectors with the activation space of a target LVLM and injects them at their corresponding generation phases via activation steering (\S\ref{sec:activation_steering}). Figure~\ref{fig:overview} shows the overall pipeline of \ours.

\subsection{Preliminaries}\label{sec:preliminaries}

\noindent\textbf{Setup.}
Let $\mathcal{D}=\{(I_i,q_i)\}_{i=1}^{N}$ denote a multimodal dataset, where $I_i$ and $q_i$ are the image and textual query, respectively.
We denote the source visual-CoT LVLM by $\mathcal{M}_s$ and the target LVLM by $\mathcal{M}_t$.
The source model $\mathcal{M}_s$ has visual-CoT capabilities, which we transfer to the target model $\mathcal{M}_t$.
Let $d_s$ and $d_t$ denote the hidden dimensions of $\mathcal{M}_s$ and $\mathcal{M}_t$, respectively.

\noindent\textbf{Visual-CoT reasoning.}
% TODO: Add citations to representative visual-CoT methods after references are finalized.
Visual-CoT, also known as \textit{``thinking with images''}, enables an LVLM to interleave textual reasoning with explicit visual grounding before producing the final answer.
Although existing visual-CoT models adopt different generation and interaction interfaces, we represent their output using the following unified format:

\begin{equation}\label{eq:visual_cot_output}
o =
\left(\left[
r_k, \mathbf{b}_k, \tilde I_k
\right]_{k=1}^{K}, a \right),
\qquad
\tilde I_k = \operatorname{Crop}(I,\mathbf b_k).
\end{equation}
This formulation represents a visual-CoT trajectory with $K$ interleaved
reasoning and grounding steps. At step $k$, $r_k$ denotes the textual
reasoning trace, $\mathbf{b}_k=[x_1,y_1,x_2,y_2]^\top$\footnote{Notation $x$ is temporarily reused for coordinates; it otherwise denotes the input.} specifies the grounded
region in the original image $I$, $\tilde I_k$ is the cropped image, and $a$ denotes the final answer.

\subsection{Disentangled visual-CoT capability extraction}
\label{sec:capability_extraction}

Visual-CoT comprises two complementary capabilities: \textbf{visual grounding} for localizing task-relevant regions and \textbf{textual reasoning} for analyzing the question and deriving the final answer.
These two capabilities operate at different layers of the model~\citep{chen2025bring}.
We therefore disentangle visual-CoT into grounding and reasoning capability vectors by constructing controlled contrastive inputs that isolate the activation changes associated with each capability, enabling them to be extracted and steered independently.

\noindent\textbf{Capability vector extraction.}
For each image--query pair $(I_i,q_i)$, we construct a positive input $x_i^{+}$ that prompts $\mathcal{M}_s$ to follow the complete visual-CoT format in Eq.~(\ref{eq:visual_cot_output}).
We then construct two negative counterparts with the image and query unchanged.
The grounding-negative input $x_{i,g}^{-}$ retains the reasoning and answering instructions but removes the bounding-box prediction and associated visual interaction.
In contrast, the reasoning-negative input $x_{i,r}^{-}$ retains the grounding and answering instructions but omits the intermediate reasoning trace. The symbols $g$ and $r$ index the grounding and reasoning capabilities, respectively.
Thus, each pair differs only in the capability of interest, reducing interference from visual content and task semantics.

We feed each contrastive input to the source model and collect the hidden state of its final input token.
For capability $c\in\{g,r\}$, let $\ell_s^c$ denote its source extraction layer, and let $\mathbf{h}_{i,c}^{+},\mathbf{h}_{i,c}^{-}\in\mathbb{R}^{d_s}$ denote the positive and negative final token activations obtained from this layer. 
Because grounding and reasoning exhibit distinct layer-wise specialization,
we select $\ell_s^g$ from the early-to-middle layers and $\ell_s^r$ from the
middle-to-late layers, following the layer-wise specialization observed in
our analysis (\S\ref{sec:intro}).
For each capability, we compute the mean activation difference over the $N$ examples:
\begin{equation}
    \mathbf{v}_{c}
    =
    \frac{1}{N}\sum_{i=1}^{N}
    \left(
        \mathbf{h}_{i,c}^{+}
        -
        \mathbf{h}_{i,c}^{-}
    \right),
    \qquad c\in\{g,r\}.
    \label{eq:capability_vector}
\end{equation}
The resulting grounding vector $\mathbf{v}_g$ represents the visual grounding capability of visual-CoT, whereas the reasoning vector $\mathbf{v}_r$ represents its reasoning capability. This disentangled construction allows the two capabilities to be transferred and controlled independently.

\subsection{Phase-aware visual-CoT activation steering}
\label{sec:activation_steering}

Because the source and target LVLMs may have different hidden dimensions,
we first map each extracted capability vector into the target activation space.
We then perform phase-aware activation steering: the reasoning vector is
injected at the target reasoning layer during reasoning-trace generation,
and the grounding vector is injected at the target grounding layer during bounding-box prediction. 
Unlike prior stage-agnostic interventions~\citep{zhan2025l2v,wang2026commandv}, our
steering is dynamically routed according to the current generation phase.

\noindent\textbf{Cross-architecture dimension alignment.}
% TODO: Cite L2V-CoT and the LMN method (Gerlach et al., 2024) after references are finalized.
Following \citet{zhan2025l2v}, we adopt the LMN~\citep{gerlach2024waveformresamplinglmnmethod} for frequency-domain resampling to map each source capability vector to the target hidden dimension.
For a capability vector $\mathbf{v}_c\in\mathbb{R}^{d_s}$, we obtain its target space representation as
\begin{equation}
    \begin{aligned}
        \overline{\mathbf{v}}_c
        &=
        \operatorname{Re}\left[
            \mathcal{F}_{d_t}^{-1}
            \left(
                \operatorname{LMN}_{d_s\rightarrow d_t}
                \left(\mathcal{F}_{d_s}(\mathbf{v}_c)\right)
            \right)
        \right],\\
        \widetilde{\mathbf{v}}_c
        &=
        \frac{\lVert\mathbf{v}_c\rVert_2}
             {\lVert\overline{\mathbf{v}}_c\rVert_2}
        \overline{\mathbf{v}}_c,
        \qquad c\in\{g,r\},
    \end{aligned}
    \label{eq:dimension_alignment}
\end{equation}
where $\mathcal{F}_{d_s}$ and $\mathcal{F}_{d_t}^{-1}$ denote the Fourier
and inverse Fourier transforms, respectively,
$\operatorname{LMN}_{d_s\rightarrow d_t}$ denotes frequency-domain
resampling from $d_s$ to $d_t$, and $\operatorname{Re}[\cdot]$ extracts
the real part of the inverse-transformed vector.
The subsequent step restores the norm of the source vector, preventing dimension alignment from changing the intervention magnitude.
This procedure produces the target-space grounding and reasoning vectors $\widetilde{\mathbf{v}}_g,\widetilde{\mathbf{v}}_r\in\mathbb{R}^{d_t}$ without updating either model.

\noindent\textbf{Phase-aware activation steering.}
Applying both capability vectors throughout generation can introduce cross-capability interference: the grounding direction may perturb textual reasoning, while the reasoning direction may interfere with bounding-box prediction.
We use a phase-aware routing strategy to synchronize activation steering with the structured visual-CoT output.
Let $z_j$ denote the generation phase at decoding step $j$, and define the active capability as
\begin{equation}
    c_j =
    \begin{cases}
        r, & z_j=\mathsf{reasoning},\\
        g, & z_j=\mathsf{grounding}.
    \end{cases}
    \label{eq:phase_policy}
\end{equation}
The routing strategy activates the reasoning vector during reasoning-trace generation and the grounding vector during bounding-box generation.
We leave the answer phase unsteered, allowing $\mathcal{M}_t$ to generate the final response based on the accumulated visual evidence and reasoning context.

At decoding step $j$ with an active capability, let $\mathbf{h}_j\in\mathbb{R}^{d_t}$ denote the current token activation at target layer $\ell_t^{c_j}$.
To preserve the scale of the original hidden state, we normalize the steered activation as
\begin{equation}
    \begin{aligned}
        \overline{\mathbf{h}}_j
        &=
        \mathbf{h}_j
        + \alpha_c\widetilde{\mathbf{v}}_{c_j},\\
        \widehat{\mathbf{h}}_j
        &=
        \frac{\lVert\mathbf{h}_j\rVert_2}
             {\lVert\overline{\mathbf{h}}_j\rVert_2}
        \overline{\mathbf{h}}_j,
    \end{aligned}
    \label{eq:phase_steering}
\end{equation}
where $\alpha_g$ and $\alpha_r$ control the grounding and reasoning steering strengths, respectively.
The intervention is applied to the current decoding position during autoregressive generation.
The target injection layers $\ell_t^g$ and $\ell_t^r$ are likewise defined
separately for grounding and reasoning, preserving their layer-wise specialization.

\section{Experiments}

\subsection{Setup}\label{sec:setup}

\noindent\textbf{Implementation details.}
We use DeepEyes-7B~\citep{zheng2026deepeyes}, a strong visual-CoT model built on Qwen2.5-VL-7B~\citep{bai2025qwen25vltechnicalreport}, as our source model.
For same-architecture transfer, we use Qwen2.5-VL-7B, the foundation model of DeepEyes-7B, as the target LVLM.
For cross-model transfer, we use Qwen2.5-VL-3B, Qwen3-VL-4B~\citep{bai2025qwen3vltechnicalreport}, InternVL2.5-8B~\citep{chen2025intervl25}, and the substantially larger Qwen3-VL-32B as target LVLMs, covering different model scales, families, layer depths, and hidden dimensions.
% We set the steering strength  $\alpha_g=\alpha_r=\alpha=0.5$ in all experiments. 
Further implementation details are provided in \S\ref{appendix-imple}, and results for Qwen2.5-VL-3B and Qwen3-VL-32B are reported in \S\ref{appendix-exp}.

\noindent\textbf{Capability vector extraction data.}
To extract the grounding and reasoning capability vectors, we use DeepEyes-7B as the source visual-CoT model and construct a 1,000-example subset from its training corpus, DeepEyes-Datasets-47k~\citep{zheng2026deepeyes}.
Following the source distribution of the original corpus, the subset contains examples of real-world visual reasoning, mathematical reasoning, and chart reasoning, covering major visual reasoning scenarios.
Details are in \S\ref{appendix-imple}.

\noindent\textbf{Benchmarks.}
Following DeepEyes, we evaluate our method on five benchmarks spanning three
visual reasoning scenarios.
\textbf{(1) High-resolution visual understanding:} V* Bench~\citep{wu2024datasetsVstar}
and HR-Bench~\citep{wang2025datasetshrbench}, which evaluate fine-grained
visual understanding under high-resolution inputs (HR-Bench includes both 4K
and 8K settings); 
\textbf{(2) General visual perception and reasoning:} MME-RealWorld-Lite~\citep{zhang2025datasetsmmerealworld},
which assesses visual perception and reasoning across diverse real-world
applications;
and \textbf{(3) Mathematical visual reasoning:} MathVista~\citep{lu2024datasetsmathvista}
and MathVerse~\citep{zhang2024datasetsmathverse}, which evaluate reasoning over
figures, diagrams, and other structured visual inputs.

\noindent\textbf{Baselines.}
We compare \ours~with the base target LVLMs without capability transfer (Vanilla), evaluated under the DeepEyes prompting setting, and three representative cross-model capability-transfer methods.
L2V-CoT~\citep{zhan2025l2v} and Command-V~\citep{wang2026commandv} transfer capabilities through activation steering, while Transport and Merge~\citep{cui2026transport} performs cross-architecture model merging.

\subsection{Main results}
\begin{table*}[t!]
\centering
\small
\setlength{\tabcolsep}{2.5pt}
\caption{
    \textbf{Main results on high-resolution visual understanding benchmarks.}
    We transfer visual-CoT capabilities from DeepEyes-7B to Qwen2.5-VL-7B, Qwen3-VL-4B and InternVL2.5-8B. Results for additional target LVLMs  (Vanilla) are reported in Table~\ref{tab:additional_main_results_hr}. We compare against the base target LVLMs, L2V-CoT, Command-V, and Transport and Merge on V* and HR-Bench. Best and second-best results are in \textbf{bold} and \underline{underlined}, respectively.
}
\label{tab:main_results_hr}
\begin{tabular}{llccccccccc}
\toprule
\multirow{2}{*}{Model}
& \multirow{2}{*}{Method}
& \multicolumn{3}{c}{V$^*$}
& \multicolumn{3}{c}{HR-Bench 4K}
& \multicolumn{3}{c}{HR-Bench 8K} \\
\cmidrule(lr){3-5}
\cmidrule(lr){6-8}
\cmidrule(lr){9-11}
& & Overall & Attr. & Spatial
& Overall & FSP & FCP
& Overall & FSP & FCP \\
\midrule

DeepEyes-7B
& Base
& 88.48 & 90.44 & 85.53
& 74.00 & 91.25 & 56.75
& 71.13 & 84.50 &57.75 \\
\midrule

\multirow{4}{*}{Qwen2.5-VL-7B}
& Vanilla
& 75.92 & 71.30 & 82.90
& 68.13 & 80.50 & 55.75
& 62.88 & 76.50& 49.25 \\
& L2V-CoT
& 84.29 & \underline{86.96} & 80.26
& \underline{71.50} & \underline{83.25} & \textbf{59.75}
& \textbf{67.50} &\textbf{81.25} & \textbf{53.75}\\
& Command-V
& \underline{86.91} & 86.09 & \underline{88.16}
& 68.75 & 81.00 & 56.50
& 64.63 & 76.50 & 52.75 \\
\rowcolor{tabhighlight}
\cellcolor{white} & Ours
& \textbf{93.72} & \textbf{91.30} & \textbf{97.37}
& \textbf{73.13} & \textbf{88.00} & \underline{58.25}
& \underline{66.88} & \underline{80.25} & \underline{53.50} \\
\midrule

\multirow{5}{*}{Qwen3-VL-4B}
& Vanilla
& 86.39 & 89.57 & 81.58
& 78.13 & 89.50 & 66.75
& 72.75 & \underline{86.75} & 58.75 \\
& L2V-CoT
& 88.48 & 88.70 & \underline{88.15}
& \textbf{79.75} & \underline{91.50} & \textbf{68.00}
& 72.50 & 83.00 & \textbf{62.00} \\
& Command-V
& \underline{89.01} & 90.44 & 86.84
& 78.38 & 89.50 & 67.25
& 72.75 & 85.50 & 60.00 \\
& Transport and Merge
& 88.48 & \underline{92.17} & 82.90
& 78.50 & 89.00 & \textbf{68.00}
& \underline{73.88} & 85.75 & \textbf{62.00} \\
\rowcolor{tabhighlight}
\cellcolor{white} & Ours
& \textbf{92.67} & \textbf{93.91} & \textbf{90.79}
& \underline{79.38} & \textbf{91.75} & 67.00
& \textbf{74.75} & \textbf{87.50} & \textbf{62.00} \\
\midrule

\multirow{5}{*}{InternVL2.5-8B}
& Vanilla
& 69.63 & 66.96 & 73.68
& 36.25 & 29.50 & 43.00
& 36.50 & 27.50 & 45.50 \\
& L2V-CoT
& \underline{70.68} & \underline{68.70} & 73.68
& \underline{37.75} & \underline{34.00} & 41.50
& 37.75 & 27.25 & \underline{48.25} \\
& Command-V
& \underline{70.68} & \underline{68.70} & 73.68
& 36.63 & 29.75 & 43.50
& \underline{37.88} & \underline{28.50} & 47.25 \\
& Transport and Merge
& \underline{70.68} & 67.83 & \underline{75.00}
& 37.63 & 28.75 & \underline{46.50}
& 37.38 & 26.50 & \underline{48.25} \\
\rowcolor{tabhighlight}
\cellcolor{white} & Ours
& \textbf{72.25} & \textbf{69.57} & \textbf{76.32}
& \textbf{50.38} & \textbf{44.75} & \textbf{56.00}
& \textbf{49.75} & \textbf{44.25} & \textbf{55.25} \\
\bottomrule
\end{tabular}
\begin{minipage}{\linewidth}
\scriptsize\textit{Note.} Transport and Merge is omitted for Qwen2.5-VL-7B; with the same backbone, full replacement with source weights recovers DeepEyes-7B (reported separately).
\end{minipage}
\end{table*}

\begin{table*}[t!]
\centering
\small
\setlength{\tabcolsep}{2.2pt}
\caption{
    \textbf{Main results on MME-RealWorld-Lite for general visual perception and reasoning.}
}
\label{tab:main_results_mme}
\begin{tabular}{llcccccccccc}
\toprule
\multirow{2}{*}{Model}
& \multirow{2}{*}{Method}
& \multirow{2}{*}{Overall}
& \multicolumn{5}{c}{Perception}
& \multicolumn{4}{c}{Reasoning} \\
\cmidrule(lr){4-8}
\cmidrule(lr){9-12}
& & & OCR & RS & DT & MO & AD
& OCR & DT & MO & AD \\
\midrule

DeepEyes-7B
& Base
& 52.27
& 90.00 & 50.67 & 89.00 & 44.51 & 34.00
& 73.00 & 70.00 & 46.00 & 35.00 \\
\midrule

\multirow{4}{*}{Qwen2.5-VL-7B}
& Vanilla
& 38.56
& 82.00 & 33.33 & \textbf{87.00} & 22.26 & 20.86
& 66.00 & \textbf{70.00} & 26.67 & 19.50 \\
& L2V-CoT
& 41.17
& \textbf{87.20} & 34.67 & 85.00 & 24.76 & 21.43
& 62.00 & 62.00 & \textbf{34.00} & \textbf{26.50} \\
& Command-V
& \underline{42.05}
& \underline{86.40} & \underline{40.67} & 82.00 & \textbf{26.65} & \underline{22.29}
& \textbf{71.00} & \underline{68.00} & 30.67 & 25.00 \\
\rowcolor{tabhighlight}
\cellcolor{white} & Ours
& \textbf{43.30}
& 84.80 & \textbf{44.67} & \underline{86.00} & \textbf{26.65} & \textbf{27.71}
& \underline{67.00} & 64.00 & \underline{31.33} & \textbf{26.50} \\
\midrule

\multirow{5}{*}{Qwen3-VL-4B}
& Vanilla
& \underline{49.66}
& 86.00 & \underline{40.67} & 80.00 & 38.87 & \textbf{39.14}
& \textbf{74.00} & \textbf{74.00} & \underline{38.67} & \underline{32.50} \\
& L2V-CoT
& 48.20
& 85.20 & 32.67 & \underline{81.00} & \underline{39.50} & 36.00
& 69.00 & \textbf{74.00} & \underline{38.67} & 32.25 \\
& Command-V
& 49.45
& \underline{86.40} & 39.33 & 80.00 & \underline{39.50} & \textbf{39.14}
& 72.00 & 68.00 & 30.67 & 25.00 \\
& Transport and Merge
& 47.84
& \underline{86.40} & 34.00 & \textbf{83.00} & 36.36 & 38.00
& 73.00 & 73.00 & \underline{38.67} & 28.75  \\
\rowcolor{tabhighlight}
\cellcolor{white} & Ours
& \textbf{50.70}
& \textbf{86.80} & \textbf{41.33} & 80.00 & \textbf{39.81} & \textbf{39.14}
&\textbf{74.00} & 70.00 & \textbf{46.67} & \textbf{34.00} \\
\midrule

\multirow{5}{*}{InternVL2.5-8B}
& Vanilla
& 44.09
& \underline{80.00} & \underline{42.00} & 65.00 & 31.66 & 36.00
& 61.00 & 50.00 & \textbf{40.67} & 29.75 \\
& L2V-CoT
& \underline{45.49}
& \underline{80.00} & 41.33 & \textbf{69.00} & 32.60 & \textbf{42.57}
& 61.00 & \underline{51.00} & 35.33 & 31.00 \\
& Command-V
& 44.14
& 79.60 & 41.33 & 64.00 & 32.92 & 36.00
& \textbf{63.00} & 48.00 & 38.00 & 30.75 \\
& Transport and Merge
& 45.23
& 76.40 & 38.00 & \textbf{69.00} & \textbf{35.11} & \underline{40.57}
& 58.00 & 46.00 & 38.67 & \textbf{33.75} \\
\rowcolor{tabhighlight}
\cellcolor{white} & Ours
& \textbf{46.17}
& \textbf{80.80} & \textbf{44.67} & \textbf{69.00} & \underline{34.48} & 36.57
& \textbf{63.00} & \textbf{52.00} & \textbf{40.67} & \underline{33.50} \\
\bottomrule
\end{tabular}
\end{table*}

\begin{table*}[t!]
\centering
\small
\setlength{\tabcolsep}{2.6pt}
\caption{
\textbf{Main results on mathematical visual reasoning benchmarks.}
}
\label{tab:main_results_math}

\begin{tabular}{llccccccccc}
\toprule
\multirow{2}{*}{Model}
& \multirow{2}{*}{Method}
& \multicolumn{3}{c}{MathVista}
& \multicolumn{6}{c}{MathVerse}
\\
\cmidrule(lr){3-5}
\cmidrule(lr){6-11}
&
& Overall & General & Math
& Overall & TD & TL & VI & VD & VO
\\
\midrule

DeepEyes-7B
& Base
& 66.5 & 69.35 & 64.07
& 43.35 & 50.76 & 44.67 &42.01 & 39.72&39.59
\\
\midrule

\multirow{4}{*}{Qwen2.5-VL-7B}
& Vanilla
& 64.9 &69.78 &\underline{60.74}
& 38.93 &45.69 &38.32 & 36.29&35.66 &\underline{38.71}
\\
& L2V-CoT
& \textbf{67.2} &\underline{70.00} &\textbf{64.81}
& \underline{40.25} & \underline{47.46}&\underline{41.12} &\underline{38.07} &\underline{38.07} &36.55
\\
& Command-V
& 64.1 &68.48 &60.37
& 39.06 & 46.19&39.09 &36.55 &35.53 &37.94
\\
\rowcolor{tabhighlight}
\cellcolor{white} &Ours
& \underline{65.6} &\textbf{71.30} &\underline{60.74}
& \textbf{42.89} & \textbf{49.37}& \textbf{45.18}&\textbf{39.85} &\textbf{40.74} &\textbf{39.34}
\\
\midrule

\multirow{5}{*}{Qwen3-VL-4B}
& Vanilla
& 66.0 &62.39 &\textbf{69.07}
& 49.72 &58.63 &50.38 &\underline{46.45} &44.92 &48.22
\\
& L2V-CoT
& \underline{66.3} & \textbf{67.39}&65.37
& 48.17 & 55.08&49.62 &44.80 &44.92 &46.45
\\
& Command-V
& 66.1 &62.61 &\textbf{69.07}
& \underline{50.63} &\underline{59.52} &\underline{51.02} &\textbf{47.34} &\textbf{46.83} &48.48
\\
& Transport and Merge
& 64.5 & 63.91&65.00
& 49.39 &58.12 &49.62 &44.80 &45.81 &\textbf{48.60}
\\
\rowcolor{tabhighlight}
\cellcolor{white} &Ours
& \textbf{66.9} &\underline{65.43} &68.15
& \textbf{50.69} &\textbf{61.29} &\textbf{51.14} &45.81 &\underline{46.57} &\textbf{48.60}
\\
\midrule

\multirow{5}{*}{InternVL2.5-8B}
& Vanilla
& 58.0 & 61.74&54.81
& \underline{35.03} &42.26 &35.79 &33.76 &\textbf{33.38} &\underline{29.95}
\\
& L2V-CoT
& \underline{60.4} &63.48 &\underline{57.78}
& 34.16 &41.50 &\underline{36.29} &\underline{34.52} &32.11 &26.40
\\
& Command-V
& 58.8 &61.74 &56.30
& 34.37 &\textbf{43.40} &35.66 &33.38 &31.98 &27.41
\\
& Transport and Merge
& 58.7 & \underline{64.13}&54.07
& 34.09 &41.50 &33.76 &33.76 &32.36 &29.06
\\
\rowcolor{tabhighlight}
\cellcolor{white} &Ours
& \textbf{61.5} &\textbf{65.22} &\textbf{58.32}
& \textbf{35.74} &\underline{42.77} &\textbf{37.44} &\textbf{34.90} &\textbf{33.38} &\textbf{30.20}
\\
\bottomrule
\end{tabular}
\end{table*}

Tables~\ref{tab:main_results_hr}, \ref{tab:main_results_mme}, and \ref{tab:main_results_math} summarize the transfer results from DeepEyes-7B to three target LVLMs spanning different model families, architectures, and scales.

\noindent\textbf{High-resolution benchmarks.}
On V* and HR-Bench, our method delivers stable improvements across most target models and evaluation settings while generally outperforming the capability-transfer baselines.
Representative gains over the corresponding base LVLMs include improvements on V* with Qwen3-VL-4B (+6.28\%) and HR-Bench 4K with InternVL2.5-8B (+14.13\%), demonstrating that the transferred capabilities remain effective across model architectures and scales.
In the same-architecture setting, our method applied to Qwen2.5-VL-7B even surpasses the source visual-CoT model DeepEyes-7B on V*.

\noindent\textbf{General visual perception and reasoning.}
On MME-RealWorld-Lite, our method improves the overall performance of all three target LVLMs and generally outperforms the capability-transfer baselines.
For example, it improves Qwen2.5-VL-7B (+4.74\%), demonstrating that the transferred visual-CoT capabilities also benefit general visual perception and reasoning.

\noindent\textbf{Mathematical reasoning benchmarks.}
On MathVista and MathVerse, our method improves all three target LVLMs.
For example, it improves InternVL2.5-8B on MathVista (+3.50\%) and Qwen2.5-VL-7B on MathVerse (+3.96\%), showing that the transferred visual-CoT capabilities generalize to mathematical visual reasoning over figures, diagrams and charts.

\subsection{Ablation studies and Visualization}\label{sec:ablation}
\noindent\textbf{Effect of capability disentanglement.}
We evaluate capability disentanglement on Qwen3-VL-4B by comparing the full method with variants that use only the grounding or reasoning vector or replace them with a unified visual-CoT vector.
As shown in Table~\ref{tab:ablation_gr}, using either capability vector alone improves performance in selected settings, while combining the two disentangled vectors achieves the best overall results on V*, MME-RealWorld-Lite, and MathVista.
The full method also consistently outperforms the unified-vector variant across the three benchmarks, with the largest improvement on V* (92.67\% vs. 88.48\%), suggesting that using disentangled grounding and reasoning vectors at their respective generation phases is more effective than applying a unified visual-CoT vector.
\begin{table}[t!]
\centering
\small
\setlength{\tabcolsep}{6pt}
\caption{
    \textbf{Ablation study of disentangled capability vectors on Qwen3-VL-4B.}
    $\widetilde{\mathbf{v}}_g$ and $\widetilde{\mathbf{v}}_r$ denote the grounding and reasoning capability vectors, respectively, while $\widetilde{\mathbf{v}}$ denotes the unified visual-CoT capability vector.
}
\label{tab:ablation_gr}

\begin{tabular}{lccccccccc}
\toprule
\multirow{2}{*}{Method}
& \multicolumn{3}{c}{V*}
& \multicolumn{3}{c}{MME-RealWorld-Lite}
& \multicolumn{3}{c}{MathVista}
\\
\cmidrule(lr){2-4}
\cmidrule(lr){5-7}
\cmidrule(lr){8-10}
& Overall & Attr. & Spatial
& Overall & Percep. & Reason.
& Overall & General & Math
\\
\midrule

Vanilla
& 86.39 & 89.57 & 81.58
& 49.66 & 52.78 & 44.80
& 66.0 & 62.39 & \textbf{69.07}
\\

\quad$+\widetilde{\mathbf{v}}_g$
& 86.91 & 89.56 & 82.90
& 49.08 & 52.70 & 43.47
& 64.4 & 61.74 & 66.67
\\

\quad$+\widetilde{\mathbf{v}}_r$
& \underline{92.15} & \textbf{95.65} & \underline{86.84}
& 50.03 & 52.78 & \underline{45.73}
& 66.3 & 63.70 & \underline{68.52}
\\

\quad$+\widetilde{\mathbf{v}}$
& 88.48 & 90.44 & 85.53
& \underline{50.44} & \textbf{53.98} & 44.93
& \underline{66.8} & \textbf{65.65} & 67.78
\\

\rowcolor{tabhighlight}
\quad$+\widetilde{\mathbf{v}}_g, +\widetilde{\mathbf{v}}_r$
& \textbf{92.67} & \underline{93.91} & \textbf{90.79}
& \textbf{50.70} & \underline{53.29} & \textbf{46.67}
& \textbf{66.9} & \underline{65.43} & 68.15
\\

\bottomrule
\end{tabular}
\end{table}

\makeatletter
\begin{figure*}[t!]
\vspace{5pt}
\noindent
\begin{minipage}[t]{0.48\textwidth}
\centering
\def\@captype{table}
\small
\setlength{\tabcolsep}{9pt}
\caption{
\textbf{Effect of phase-aware routing on Qwen3-VL-4B.}
``w/o Routing'' applies both $\widetilde{\mathbf{v}}_g$ and $\widetilde{\mathbf{v}}_r$ at every decoding step.
}
\label{tab:ablation_routing}
\begin{tabular}{lccc}
\toprule
\multirow{2}{*}{Method}
& \multicolumn{3}{c}{V*}
\\
\cmidrule(lr){2-4}
& Overall & Attr. & Spatial
\\
\midrule
Vanilla
& \underline{86.39} & 89.57 & \underline{81.58}
\\
w/o Routing
& 85.34 & \underline{91.30} & 76.32
\\
\rowcolor{tabhighlight}
Ours
& \textbf{92.67} & \textbf{93.91} & \textbf{90.79}
\\
\bottomrule
\end{tabular}
% \par\vspace{0.8em}
\def\@captype{figure}
\includegraphics[width=\linewidth]{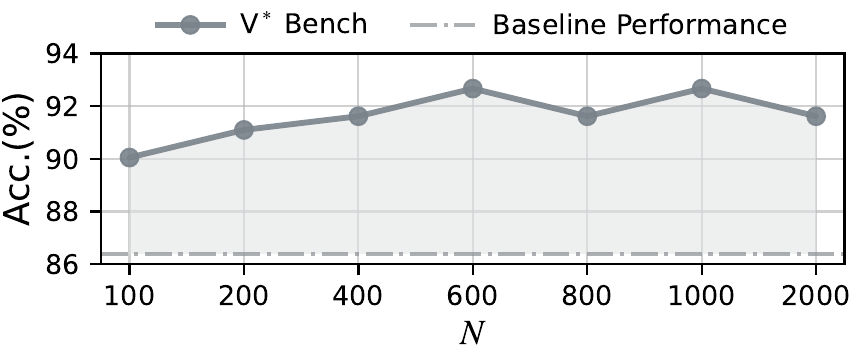}
\caption{
\textbf{Effect of $N$ on Qwen3-VL-4B.}
}
\label{fig:ablation_n}
\end{minipage}\hfill
\begin{minipage}[t]{0.495\textwidth}
\vspace{0pt}
\centering
\def\@captype{figure}
\includegraphics[width=\linewidth]{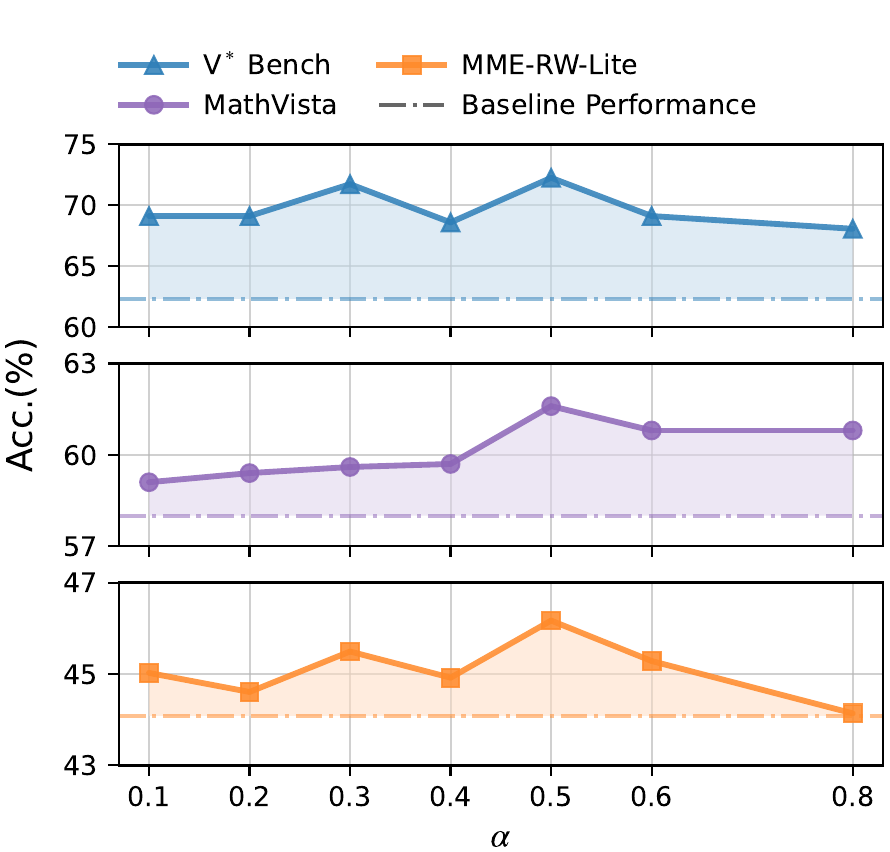}
\caption{
\textbf{Effect of $\alpha$ on InternVL-2.5-8B.}
}
\label{fig:alpha_ablation_internvl}
\end{minipage}
\end{figure*}
\makeatother

\noindent\textbf{Effect of phase-aware routing.}
We compare our phase-aware routing with a variant that applies both capability vectors throughout generation.
As shown in Table~\ref{tab:ablation_routing}, removing phase-aware routing reduces V* accuracy from 92.67\% to 85.34\%, even below the base model's 86.39\%.
This result shows that interventions irrelevant to the current generation phase can offset the benefits of the two capability vectors, while routing each vector to its corresponding phase effectively combines them.

\noindent\textbf{Effect of steering strength.}
We evaluate the effect of different steering strengths $\alpha$ on performance using InternVL2.5-8B. The results are shown in Figure~\ref{fig:alpha_ablation_internvl}. \ours~maintains stable improvements over a range of $\alpha$ values, demonstrating robustness to the steering strength, while $\alpha=0.5$ yields the strongest and most consistent performance across the three benchmarks.

\begin{figure*}[t]
    \centering
\includegraphics[width=\textwidth]{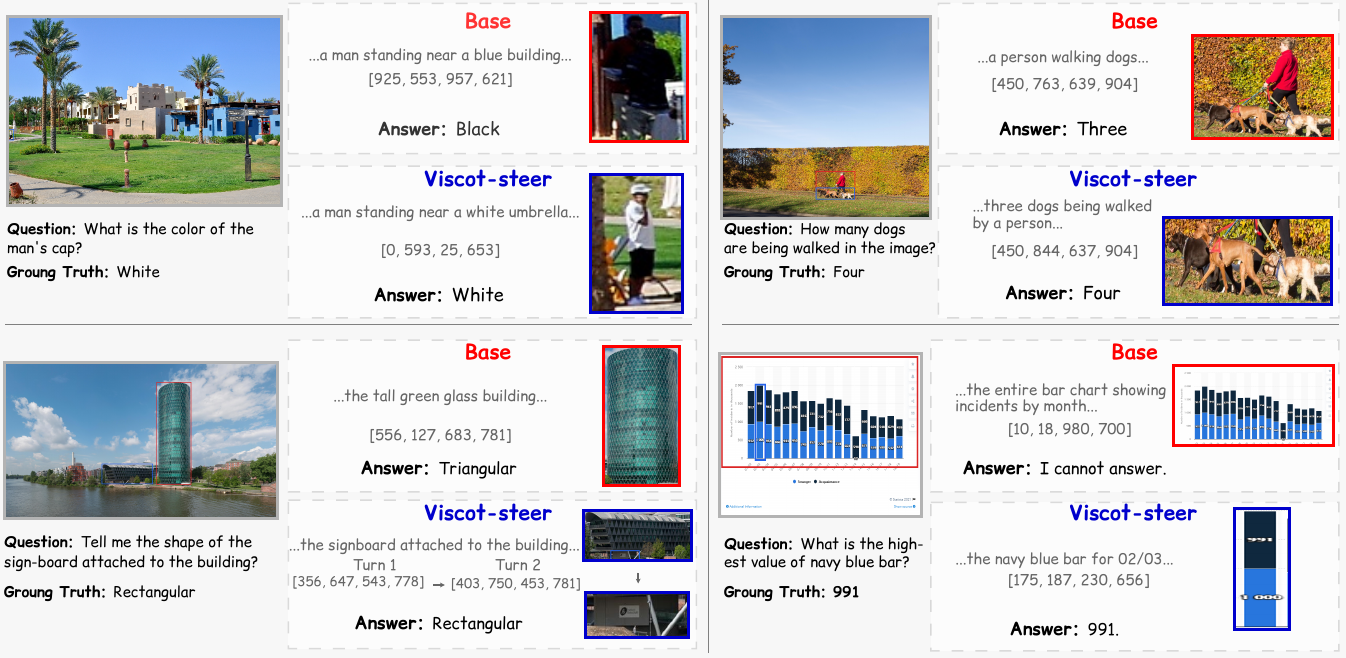}
    \caption{Visualization of the responses generated by the base LVLM and \ours.}
    \label{fig:visualization}
\end{figure*}
\noindent\textbf{Effect of extraction data size.}
We investigate how the number of examples $N$ used for capability-vector extraction affects performance. The results are shown in Figure~\ref{fig:ablation_n}.
With only 100 examples, \ours~already achieves 90.05\%, substantially outperforming the base model's 86.39\%.
Performance generally improves as $N$ increases, reaching 92.67\% at $N=600$, after which further increases yield no consistent gains.
This suggests that our method requires only a small extraction set to capture general and robust visual-CoT representations.

\makeatletter
\begin{figure}[t!]
\noindent
\begin{minipage}[t]{0.52\textwidth}
\vspace{0pt}
\def\@captype{table}
\centering
\small
\setlength{\tabcolsep}{1.9pt}
\caption{
    \textbf{Transfer results} with GRIT as the source Visual-CoT model.
}
\vspace{3pt}
\label{tab:grit_source}
\begin{tabular}{llcc}
\toprule
Model & Method & VSR & TallyQA \\
\midrule

GRIT-3B
& Base
& 75.09 & 47.45 \\
\midrule

\multirow{5}{*}{Qwen3-VL-4B}
& Vanilla
& 55.96 & 51.53 \\
& L2V-CoT
& \underline{61.73}&52.34\\
& Command-V
& 58.84&\underline{53.56} \\
& Transport and Merge
& 54.87 & 50.71 \\
\rowcolor{tabhighlight}
\cellcolor{white} & Ours
& \textbf{64.26} & \textbf{54.58} \\
\midrule

\multirow{5}{*}{InternVL3-2B}
& Vanilla
& 33.57 &38.90 \\
& L2V-CoT
& \textbf{48.01}&40.33 \\
& Command-V
&36.82 &\underline{40.94} \\
& Transport and Merge
& 42.60 & 38.49 \\
\rowcolor{tabhighlight}
\cellcolor{white} & Ours
& \underline{46.57} & \textbf{42.36} \\
\bottomrule
\end{tabular}

\end{minipage}\hfill
\begin{minipage}[t]{0.48\textwidth}
\vspace{0pt}
\centering
\def\@captype{figure}
\vspace{7pt}
\includegraphics[width=\linewidth]{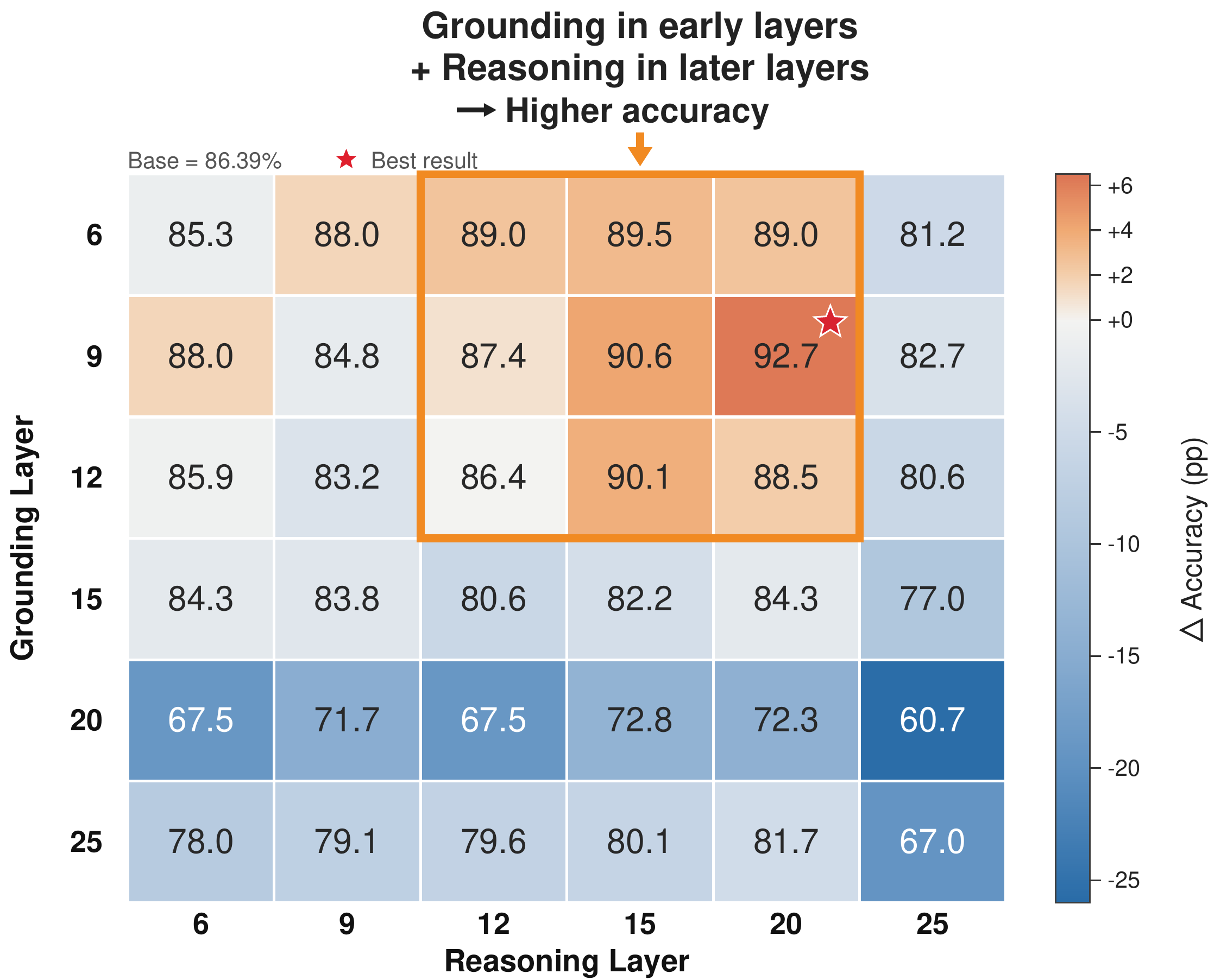}
\vspace{-19pt}
\caption{\textbf{Effect of steering layer}}
\label{fig:ablation_layer}
\end{minipage}
\end{figure}
\makeatother

\noindent\textbf{Effect of steering layers.}
We sweep the grounding and reasoning steering layers on Qwen3-VL-4B using V*.
As shown in Figure~\ref{fig:ablation_layer}, earlier or intermediate grounding layers generally perform best when paired with later reasoning layers, while injecting the grounding vector into very late layers substantially degrades performance.
The optimal combination, grounding at layer 9 and reasoning at layer 20, achieves 92.67\% (+6.28\%).
These results support steering the two capabilities at distinct, capability-specific layers and are consistent with grounding preceding higher-level reasoning.

\noindent\textbf{Visualization.}
Figure~\ref{fig:visualization} presents representative examples comparing the outputs of the base LVLM and \ours.
By steering the model toward task-relevant regions, \ours~corrects the erroneous output trajectories of the base model and produces the correct answers.

\subsection{Transfer from an alternative source model}
To evaluate transfer from a different source visual-CoT model, we use GRIT~\citep{fan2025grit}, built on Qwen2.5-VL-3B, as the source model. We use Qwen3-VL-4B, and InternVL3-2B~\citep{zhu2025internvl3exploringadvancedtraining} as target LVLMs and evaluate on VSR~\citep{liu-etal-2023-visual} and TallyQA~\citep{acharya2019tallyqa}. We follow the evaluation protocol of GRIT~\citep{fan2025grit}.

As shown in Table~\ref{tab:grit_source}, \ours~improves all target LVLMs on both benchmarks. For example, it improves Qwen3-VL-4B on VSR (+8.30\%) and TallyQA (+3.05\%). These results further demonstrate that \ours~can transfer visual-CoT capabilities from different source models.

\section{Conclusion}
In this paper, we proposed~\ours, a training-free framework for transferring visual-CoT capabilities across heterogeneous LVLMs. \ours~disentangles visual-CoT into grounding and reasoning capability vectors, aligns them with the target activation space, and steers each capability at its corresponding model layer and generation phase. Extensive experiments across five target LVLMs and five visual reasoning benchmarks demonstrate that \ours~outperforms representative capability-transfer baselines in most model--benchmark settings, while experiments with different source visual-CoT models further validate its generalizability.

\subsection*{AI use statement}
In this work, we used generative AI tools, primarily OpenAI’s GPT-5.6 Sol and GPT-5.5, solely for language polishing, translation, and manuscript formatting. We did not use generative AI tools for other research tasks, including developing research ideas, designing or evaluating methodology or experiments, implementing methods, generating synthetic data, processing research data, or interpreting results. All AI-assisted text and formatting were manually reviewed by the authors, who take full responsibility for the final manuscript and research artifacts.

\subsection*{Ethics statement}
This work adheres to the ICLR Code of Ethics. In this study, no human subjects or animal experimentation was involved. All datasets used, including V*, HR-Bench, and MathVista, among others, were sourced in compliance with usage guidelines, ensuring no violation of privacy. We have taken care to avoid any biases or discriminatory outcomes in our research process. No personally identifiable information was used, and no experiments were conducted that could raise privacy or security concerns. We are committed to transparency and integrity throughout the research process.
\subsection*{Reproducibility statement}
Our experiments use publicly available backbone models and benchmarks, including Qwen2.5-VL, Qwen3-VL, InternVL, V*, HR-Bench, and MathVista. The method section describes capability-vector extraction, cross-model dimension alignment, and phase-aware activation steering. Appendix~\ref{appendix-imple} provides implementation details, including contrastive prompts, the composition of the extraction dataset, hyperparameter settings for the main experiments, and inference and evaluation settings. Together with the publicly available models and datasets, these details provide sufficient information to reproduce our method and independently verify the reported results.

% \bibliography{iclr2026_conference}
\bibliography{main}

@inproceedings{
wang2026commandv,
title={Command-V: Training-Free Representation Finetuning Transfer},
author={Barry Wang and Avi Schwarzschild and Alexander Robey and Ali Payani and Charles Fleming and Mingjie Sun and Daphne Ippolito},
booktitle={ICLR},
year={2026},
}

@inproceedings{zhan2025l2v,
  title={L2V-CoT: Cross-Modal Transfer of Chain-of-Thought Reasoning via Latent Intervention},
  author={Zhan, Yuliang and Tang, Xinyu and Wan, Han and Li, Jian and Wen, Ji-Rong and Sun, Hao},
  booktitle={AAAI},
  year={2026}
}

@misc{chen2025intervl25,
      title={Expanding Performance Boundaries of Open-Source Multimodal Models with Model, Data, and Test-Time Scaling}, 
      author={Zhe Chen and Weiyun Wang and Yue Cao and Yangzhou Liu and Zhangwei Gao and Erfei Cui and Jinguo Zhu and Shenglong Ye and Hao Tian and Zhaoyang Liu and Lixin Gu and Xuehui Wang and Qingyun Li and Yiming Ren and Zixuan Chen and Jiapeng Luo and Jiahao Wang and Tan Jiang and Bo Wang and Conghui He and Botian Shi and Xingcheng Zhang and Han Lv and Yi Wang and Wenqi Shao and Pei Chu and Zhongying Tu and Tong He and Zhiyong Wu and Huipeng Deng and Jiaye Ge and Kai Chen and Kaipeng Zhang and Limin Wang and Min Dou and Lewei Lu and Xizhou Zhu and Tong Lu and Dahua Lin and Yu Qiao and Jifeng Dai and Wenhai Wang},
      year={2024},
      eprint={2412.05271},
      archivePrefix={arXiv},
      primaryClass={cs.CV}, 
}

@misc{bai2025qwen25vltechnicalreport,
      title={Qwen2.5-VL Technical Report}, 
      author={Shuai Bai and Keqin Chen and Xuejing Liu and Jialin Wang and Wenbin Ge and Sibo Song and Kai Dang and Peng Wang and Shijie Wang and Jun Tang and Humen Zhong and Yuanzhi Zhu and Mingkun Yang and Zhaohai Li and Jianqiang Wan and Pengfei Wang and Wei Ding and Zheren Fu and Yiheng Xu and Jiabo Ye and Xi Zhang and Tianbao Xie and Zesen Cheng and Hang Zhang and Zhibo Yang and Haiyang Xu and Junyang Lin},
      year={2025},
      eprint={2502.13923},
      archivePrefix={arXiv},
      primaryClass={cs.CV},
}

@misc{bai2025qwen3vltechnicalreport,
      title={Qwen3-VL Technical Report}, 
      author={Shuai Bai and Yuxuan Cai and Ruizhe Chen and Keqin Chen and Xionghui Chen and Zesen Cheng and Lianghao Deng and Wei Ding and Chang Gao and Chunjiang Ge and Wenbin Ge and Zhifang Guo and Qidong Huang and Jie Huang and Fei Huang and Binyuan Hui and Shutong Jiang and Zhaohai Li and Mingsheng Li and Mei Li and Kaixin Li and Zicheng Lin and Junyang Lin and Xuejing Liu and Jiawei Liu and Chenglong Liu and Yang Liu and Dayiheng Liu and Shixuan Liu and Dunjie Lu and Ruilin Luo and Chenxu Lv and Rui Men and Lingchen Meng and Xuancheng Ren and Xingzhang Ren and Sibo Song and Yuchong Sun and Jun Tang and Jianhong Tu and Jianqiang Wan and Peng Wang and Pengfei Wang and Qiuyue Wang and Yuxuan Wang and Tianbao Xie and Yiheng Xu and Haiyang Xu and Jin Xu and Zhibo Yang and Mingkun Yang and Jianxin Yang and An Yang and Bowen Yu and Fei Zhang and Hang Zhang and Xi Zhang and Bo Zheng and Humen Zhong and Jingren Zhou and Fan Zhou and Jing Zhou and Yuanzhi Zhu and Ke Zhu},
      year={2025},
      eprint={2511.21631},
      archivePrefix={arXiv},
      primaryClass={cs.CV},
}

@misc{zhu2025internvl3exploringadvancedtraining,
      title={InternVL3: Exploring Advanced Training and Test-Time Recipes for Open-Source Multimodal Models}, 
      author={Jinguo Zhu and Weiyun Wang and Zhe Chen and Zhaoyang Liu and Shenglong Ye and Lixin Gu and Hao Tian and Yuchen Duan and Weijie Su and Jie Shao and Zhangwei Gao and Erfei Cui and Xuehui Wang and Yue Cao and Yangzhou Liu and Xingguang Wei and Hongjie Zhang and Haomin Wang and Weiye Xu and Hao Li and Jiahao Wang and Nianchen Deng and Songze Li and Yinan He and Tan Jiang and Jiapeng Luo and Yi Wang and Conghui He and Botian Shi and Xingcheng Zhang and Wenqi Shao and Junjun He and Yingtong Xiong and Wenwen Qu and Peng Sun and Penglong Jiao and Han Lv and Lijun Wu and Kaipeng Zhang and Huipeng Deng and Jiaye Ge and Kai Chen and Limin Wang and Min Dou and Lewei Lu and Xizhou Zhu and Tong Lu and Dahua Lin and Yu Qiao and Jifeng Dai and Wenhai Wang},
      year={2025},
      eprint={2504.10479},
      archivePrefix={arXiv},
      primaryClass={cs.CV},
}

@inproceedings{
zheng2026deepeyes,
title={DeepEyes: Incentivizing ''Thinking with Images'' via Reinforcement Learning},
author={Ziwei Zheng and Michael Yang and Jack Hong and Chenxiao Zhao and Guohai Xu and Le Yang and Chao Shen and Xing Yu},
booktitle={ICLR},
year={2026},
}

@inproceedings{zhao2025uvcot,
      title={Unsupervised Visual Chain-of-Thought Reasoning via Preference Optimization}, 
      author={Kesen Zhao and Beier Zhu and Qianru Sun and Hanwang Zhang},
      year={2025},
      booktitle={ICCV}, 
}

@inproceedings{lai2025mini-o3,
  title={Mini-o3: Scaling Up Reasoning Patterns and Interaction Turns for Visual Search},
  author={Lai, Xin and Li, Junyi and Li, Wei and Liu, Tao and Li, Tianjian and Zhao, Hengshuang},
  booktitle={ICLR}, 
  year={2026}
}

@inproceedings{
zhang2026thyme,
title={Thyme: Think Beyond Images},
author={YiFan Zhang and Xingyu Lu and Shukang Yin and Chaoyou Fu and Wei Chen and Xiao Hu and Bin Wen and Kaiyu Jiang and Changyi Liu and Tianke Zhang and Haonan Fan and Kaibing Chen and Jiankang Chen and Haojie Ding and Kaiyu Tang and Zhang Zhang and Liang Wang and Fan Yang and Tingting Gao and Guorui Zhou},
booktitle={ICLR},
year={2026}
}

@inproceedings{pixelreasoner,
      title={Pixel Reasoner: Incentivizing Pixel-Space Reasoning with Curiosity-Driven Reinforcement Learning},
      author = {Su, Alex and Wang, Haozhe and Ren, Weiming and Lin, Fangzhen and Chen, Wenhu},
      booktitle={NeurIPS},
      year={2025}
}

@inproceedings{fan2025grit,
      title={GRIT: Teaching MLLMs to Think with Images}, 
      author={Yue Fan and Xuehai He and Diji Yang and Kaizhi Zheng and Ching-Chen Kuo and Yuting Zheng and Sravana Jyothi Narayanaraju and Xinze Guan and Xin Eric Wang},
      booktitle={NeurIPS},
      year={2025}
}

@inproceedings{wang2026vgr,
  title={Vgr: Visual grounded reasoning},
  author={Wang, Jiacong and Kang, Zijian and Wang, Haochen and Liang, Xiao and Wang, Ya and Li, Jiawen and Wu, Bohong and Ran, Jiao and Jiang, Haiyong and Feng, Chao and others},
  booktitle={ICLR},
  year={2026}
}

@article{turner2023steering,
  title={Steering language models with activation engineering},
  author={Turner, Alexander Matt and Thiergart, Lisa and Leech, Gavin and Udell, David and Vazquez, Juan J and Mini, Ulisse and MacDiarmid, Monte},
  journal={arXiv preprint arXiv:2308.10248},
  year={2023}
}

@inproceedings{li2024inferenceintervent,
  title={Inference-time intervention: Eliciting truthful answers from a language model},
  author={Li, Kenneth and Patel, Oam and Vi{\'e}gas, Fernanda and Pfister, Hanspeter and Wattenberg, Martin},
  booktitle={NeurIPS},
  year={2023}
}

@inproceedings{rimsky-etal-2024-steering,
    title = "Steering Llama 2 via Contrastive Activation Addition",
    author = "Rimsky, Nina  and
      Gabrieli, Nick  and
      Schulz, Julian  and
      Tong, Meg  and
      Hubinger, Evan  and
      Turner, Alexander",
    booktitle = "ACL",
    year = "2024",
}

@inproceedings{
arditi2024refusal,
title={Refusal in Language Models Is Mediated by a Single Direction},
author={Andy Arditi and Oscar Balcells Obeso and Aaquib Syed and Daniel Paleka and Nina Rimsky and Wes Gurnee and Neel Nanda},
booktitle={NeurIPS},
year={2024},
}

@inproceedings{tang-etal-2025-unlocking,
    title = "Unlocking General Long Chain-of-Thought Reasoning Capabilities of Large Language Models via Representation Engineering",
    author = "Tang, Xinyu  and
      Wang, Xiaolei  and
      Lv, Zhihao  and
      Min, Yingqian  and
      Zhao, Wayne Xin  and
      Hu, Binbin  and
      Liu, Ziqi  and
      Zhang, Zhiqiang",
    booktitle = "ACL",
    year = "2025",
}

@inproceedings{
zhao2025activation,
title={Activation Control for Efficiently Eliciting Long Chain-of-thought Ability of Language Models},
author={Zekai Zhao and Qi Liu and Kun Zhou and Zihan Liu and Yifei Shao and Zhiting Hu and Biwei Huang},
booktitle={NeurIPS},
year={2025},

}

@inproceedings{gan-etal-2026-textual,
    title = "Textual Steering Vectors Can Improve Visual Understanding in Multimodal Large Language Models",
    author = "Gan, Woody Haosheng  and
      Fu, Deqing  and
      Asilis, Julian  and
      Liu, Ollie  and
      Sharan, Vatsal  and
      Jia, Robin  and
      Neiswanger, Willie",
    booktitle = "ACL",
    year = "2026"
}

@inproceedings{
chen2025bring,
title={Bring Reason to Vision: Understanding Perception and Reasoning through Model Merging},
author={Shiqi Chen and Jinghan Zhang and Tongyao Zhu and Wei Liu and Siyang Gao and Miao Xiong and Manling Li and Junxian He},
booktitle={ICML},
year={2025}
}

@inproceedings{
cui2026transport,
title={Transport and Merge: Cross-Architecture Merging for Large Language Models},
author={Chenhang Cui and Binyun Yang and Fei Shen and Yuxin Chen and Jingnan Zheng and Xiang Wang and An Zhang and Tat-Seng Chua},
booktitle={ICML},
year={2026},
}

@misc{gerlach2024waveformresamplinglmnmethod,
      title={Waveform resampling with LMN method}, 
      author={Lino Gerlach and Wenqiang Gu and Nitish Nayak and Xin Qian and Brett Viren},
      year={2024},
      eprint={2407.21750},
      archivePrefix={arXiv},
      primaryClass={hep-ex},
}

@inproceedings{wu2024datasetsVstar,
  title={V*: Guided Visual Search as a Core Mechanism in Multimodal LLMs},
  author={Penghao Wu and Saining Xie},
  booktitle={CVPR},
  year={2024}
}

@inproceedings{wang2025datasetshrbench,
  title={Divide, conquer and combine: A training-free framework for high-resolution image perception in multimodal large language models},
  author={Wang, Wenbin and Ding, Liang and Zeng, Minyan and Zhou, Xiabin and Shen, Li and Luo, Yong and Yu, Wei and Tao, Dacheng},
  booktitle={AAAI},
  year={2025}
}

@inproceedings{
zhang2025datasetsmmerealworld,
title={{MME}-RealWorld: Could Your Multimodal {LLM} Challenge High-Resolution Real-World Scenarios that are Difficult for Humans?},
author={YiFan Zhang and Huanyu Zhang and Haochen Tian and Chaoyou Fu and Shuangqing Zhang and Junfei Wu and Feng Li and Kun Wang and Qingsong Wen and Zhang Zhang and Liang Wang and Rong Jin},
booktitle={ICLR},
year={2025},
}

@inproceedings{lu2024datasetsmathvista,
  title={MathVista: Evaluating Mathematical Reasoning of Foundation Models in Visual Contexts},
  author={Lu, Pan and Bansal, Hritik and Xia, Tony and Liu, Jiacheng and Li, Chunyuan and Hajishirzi, Hannaneh and Cheng, Hao and Chang, Kai-Wei and Galley, Michel and Gao, Jianfeng},
  booktitle={ICLR},
  year={2024}
}

@inproceedings{zhang2024datasetsmathverse,
  title={MathVerse: Does Your Multi-modal LLM Truly See the Diagrams in Visual Math Problems?},
  author={Zhang, Renrui and Jiang, Dongzhi and Zhang, Yichi and Lin, Haokun and Guo, Ziyu and Qiu, Pengshuo and Zhou, Aojun and Lu, Pan and Chang, Kai-Wei and Qiao, Yu and Gao, Peng and Li, Hongsheng },
  booktitle={ECCV},
  year={2024}
}

@article{liu-etal-2023-visual,
    title = "Visual Spatial Reasoning",
    author = "Liu, Fangyu  and
      Emerson, Guy  and
      Collier, Nigel",
    journal= "TACL",
    year = "2023",
}

@inproceedings{acharya2019tallyqa,
  title={TallyQA: Answering Complex Counting Questions},
  author={Acharya, Manoj and Kafle, Kushal and Kanan, Christopher},
  booktitle={AAAI},
  year={2019}
}

@inproceedings{wu-etal-2025-sharp,
    title = "{SHARP}: Steering Hallucination in {LVLM}s via Representation Engineering",
    author = "Wu, Junfei  and
      Ding, Yue  and
      Liu, Guofan  and
      Xia, Tianze  and
      Huang, Ziyue  and
      Sui, Dianbo  and
      Liu, Qiang  and
      Wu, Shu  and
      Wang, Liang  and
      Tan, Tieniu",
    booktitle = "EMNLP",
    year = "2025",
}

@misc{zhao2026nvcot,
      title={Thinking with Images as Continuous Actions: Numerical Visual Chain-of-Thought}, 
      author={Kesen Zhao and Beier Zhu and Junbao Zhou and Xingyu Zhu and Zhongqi Yue and Hanwang Zhang},
      year={2026},
      eprint={2602.23959},
      archivePrefix={arXiv},
      primaryClass={cs.CV},
}

@inproceedings{shao2024visual,
      title={Visual CoT: Advancing Multi-Modal Language Models with a Comprehensive Dataset and Benchmark for Chain-of-Thought Reasoning}, 
      author={Hao Shao and Shengju Qian and Han Xiao and Guanglu Song and Zhuofan Zong and Letian Wang and Yu Liu and Hongsheng Li},
      year={2024},
      booktitle={NeurIPS}
}

@inproceedings{li-etal-2024-multimodal-arxiv,
    title = "Multimodal {A}r{X}iv: A Dataset for Improving Scientific Comprehension of Large Vision-Language Models",
    author = "Li, Lei  and
      Wang, Yuqi  and
      Xu, Runxin  and
      Wang, Peiyi  and
      Feng, Xiachong  and
      Kong, Lingpeng  and
      Liu, Qi",
    editor = "Ku, Lun-Wei  and
      Martins, Andre  and
      Srikumar, Vivek",
    booktitle = "ACL",
    year = "2024",
}

@inproceedings{wang2025sota,
  title={SoTA with Less: MCTS-Guided Sample Selection for Data-Efficient Visual Reasoning Self-Improvement},
  author={Wang, Xiyao and Yang, Zhengyuan and Feng, Chao and Lu, Hongjin and Li, Linjie and Lin, Chung-Ching and Lin, Kevin and Huang, Furong and Wang, Lijuan},
  booktitle={NeurIPS},
  year={2025}
}

@inproceedings{hu2024visualsketchpad,
 author = {Hu, Yushi and Shi, Weijia and Fu, Xingyu and Roth, Dan and Ostendorf, Mari and Zettlemoyer, Luke and Smith, Noah and Krishna, Ranjay},
 booktitle = {NeurIPS},
 title = {Visual Sketchpad: Sketching as a Visual Chain of Thought for Multimodal Language Models},
 year = {2024}
}

@article{Schönemann_1966, 
title={A Generalized Solution of the Orthogonal Procrustes Problem},
journal={Psychometrika}, 
author={Schönemann, Peter H.}, 
year={1966}}

@misc{tang2025enhancingcrosstasktransferlarge,
      title={Enhancing Cross-task Transfer of Large Language Models via Activation Steering}, 
      author={Xinyu Tang and Zhihao Lv and Xiaoxue Cheng and Junyi Li and Wayne Xin Zhao and Zujie Wen and Zhiqiang Zhang and Jun Zhou},
      year={2025},
      eprint={2507.13236},
      archivePrefix={arXiv},
      primaryClass={cs.CL},
}

@inproceedings{huang-etal-2025-cross,
    title = "Cross-model Transferability among Large Language Models on the Platonic Representations of Concepts",
    author = "Huang, Youcheng  and
      Huang, Chen  and
      Feng, Duanyu  and
      Lei, Wenqiang  and
      Lv, Jiancheng",
    editor = "Che, Wanxiang  and
      Nabende, Joyce  and
      Shutova, Ekaterina  and
      Pilehvar, Mohammad Taher",
    booktitle = "ACL",
    year = "2025",
}

@inproceedings{wang-etal-2024-inferaligner,
    title = "{I}nfer{A}ligner: Inference-Time Alignment for Harmlessness through Cross-Model Guidance",
    author = "Wang, Pengyu  and
      Zhang, Dong  and
      Li, Linyang  and
      Tan, Chenkun  and
      Wang, Xinghao  and
      Zhang, Mozhi  and
      Ren, Ke  and
      Jiang, Botian  and
      Qiu, Xipeng",
    booktitle = "EMNLP",
    year = "2024",
}
\bibliographystyle{iclr2026_conference}

\newpage

\appendix
\section{Implementation Details}
\label{appendix-imple}

\subsection{Prompts}
\label{appendix-prompts}
We use three user prompts to extract contrastive capability vectors from DeepEyes-7B~\citep{zheng2026deepeyes}, the source model in our main experiments. The official visual-CoT prompt serves as the positive input for both capabilities. The no-thinking prompt removes intermediate reasoning while retaining tool use, whereas the no-grounding prompt disables tool use while retaining reasoning. Each contrastive pair uses the same image and question. The prompts are below:

\begin{promptbox}{Visual-CoT prompt}
\textbf{System prompt:} Official DeepEyes system prompt~\citep{zheng2026deepeyes}.
\par\smallskip
\textbf{User prompt:} Think first, call \texttt{image\_zoom\_in\_tool} if needed, then answer.
Format strictly as:
\par\smallskip
\texttt{\textless think\textgreater...\textless/think\textgreater}
\texttt{\textless tool\_call\textgreater...\textless/tool\_call\textgreater} (if tools are needed)
\texttt{\textless answer\textgreater...\textless/answer\textgreater}
\end{promptbox}

\begin{promptbox}{No-thinking prompt}
\textbf{System prompt:} Official DeepEyes system prompt~\citep{zheng2026deepeyes}.
\par\smallskip
\textbf{User prompt:} Call \texttt{image\_zoom\_in\_tool} if close visual inspection is needed; otherwise answer directly.
Do not write any thinking, rationale, analysis, or \texttt{\textless think\textgreater} tags.
Output exactly one of these formats:
\par\smallskip
\texttt{\textless tool\_call\textgreater...\textless/tool\_call\textgreater}
or
\texttt{\textless answer\textgreater...\textless/answer\textgreater}
\par\smallskip
If answering, put only the selected option letter and answer text inside \texttt{\textless answer\textgreater...\textless/answer\textgreater}.
\end{promptbox}

\begin{promptbox}{No-grounding prompt}
\textbf{User prompt:} Think first, then answer. Do not call tools.
Format strictly as:
\par\smallskip
\texttt{\textless think\textgreater...\textless/think\textgreater}
\texttt{\textless answer\textgreater...\textless/answer\textgreater}
\end{promptbox}

\subsection{Hyperparameter Settings}
The source extraction layers, target steering layers, and steering strengths are the key hyperparameters for activation-based capability transfer. To ensure a fair comparison, we tune the key hyperparameters of \ours~and all baselines using the same training split for each benchmark. For \ours, we fix the source grounding and reasoning extraction layers and tune the corresponding target steering layers and shared steering strength. For L2V-CoT, we tune the source extraction layer, target injection layer, and injection strength. For Command-V, we tune the injection layer and injection strength. The detailed settings are reported in Tables~\ref{tab:hyperparameters_ours}, \ref{tab:hyperparameters_l2vcot}, and~\ref{tab:hyperparameters_commandv}. For Transport and Merge~\citep{cui2026transport}, we set $\alpha=0.05$, \texttt{neuron-topk}$=128$, and \texttt{layer-topk}$=3$.

\begin{table*}[t!]
\centering
\small
\setlength{\tabcolsep}{3.1pt}
\caption{
    \textbf{Hyperparameter settings of \ours.}
    $\ell_t^g$ and $\ell_t^r$ denote the target grounding and reasoning steering layers, respectively. $\alpha$ denotes the steering strength, where $\alpha=\alpha_g=\alpha_r$. The source extraction layers are fixed to $\ell_s^g=12$ and $\ell_s^r=22$ for all settings. MME-RW-Lite denotes MME-RealWorld-Lite.
}
\label{tab:hyperparameters_ours}
\begin{tabular}{l*{3}{>{\centering\arraybackslash}p{0.040\textwidth}}*{3}{>{\centering\arraybackslash}p{0.040\textwidth}}*{3}{>{\centering\arraybackslash}p{0.040\textwidth}}*{3}{>{\centering\arraybackslash}p{0.040\textwidth}}*{3}{>{\centering\arraybackslash}p{0.040\textwidth}}}
\toprule
\multirow{2}{*}{Model}
& \multicolumn{3}{c}{V*}
& \multicolumn{3}{c}{HR-Bench}
& \multicolumn{3}{c}{MME-RW-Lite}
& \multicolumn{3}{c}{MathVista}
& \multicolumn{3}{c}{MathVerse} \\
\cmidrule(lr){2-4}
\cmidrule(lr){5-7}
\cmidrule(lr){8-10}
\cmidrule(lr){11-13}
\cmidrule(lr){14-16}
& $\ell_t^g$ & $\ell_t^r$ & $\alpha$
& $\ell_t^g$ & $\ell_t^r$ & $\alpha$
& $\ell_t^g$ & $\ell_t^r$ & $\alpha$
& $\ell_t^g$ & $\ell_t^r$ & $\alpha$
& $\ell_t^g$ & $\ell_t^r$ & $\alpha$ \\
\midrule
Qwen2.5-VL-7B
& 12 & 22 & 1.0
& 12 & 22 & 1.0
& 12 & 22 & 1.0
& 12 & 25 & 1.0
& 12 & 25 & 1.0 \\
Qwen3-VL-4B
& 9 & 20 & 0.5
& 6 & 25 & 0.5
& 9 & 20 & 0.5
& 15 & 23 & 0.5
& 15 & 23 & 0.5 \\
InternVL2.5-8B
& 12 & 23 & 0.5
& 12 & 23 & 0.5
& 15 & 27 & 0.5
& 16 & 22 & 0.5
& 16 & 22 & 0.5 \\
\bottomrule
\end{tabular}
\end{table*}

\begin{table*}[t!]
\centering
\small
\setlength{\tabcolsep}{1.8pt}
\caption{
    \textbf{Hyperparameter settings of L2V-CoT~\citep{zhan2025l2v}.}
    $\ell_s$ and $\ell_t$ denote the source extraction and target injection layers, respectively, and $\alpha$ denotes the injection strength.
}
\label{tab:hyperparameters_l2vcot}
\begin{tabular}{l*{3}{>{\centering\arraybackslash}p{0.037\textwidth}}*{3}{>{\centering\arraybackslash}p{0.037\textwidth}}*{3}{>{\centering\arraybackslash}p{0.037\textwidth}}*{3}{>{\centering\arraybackslash}p{0.040\textwidth}}*{3}{>{\centering\arraybackslash}p{0.035\textwidth}}*{3}{>{\centering\arraybackslash}p{0.035\textwidth}}}
\toprule
\multirow{2}{*}{Target model}
& \multicolumn{3}{c}{V*}
& \multicolumn{3}{c}{HR-Bench 4K}
& \multicolumn{3}{c}{HR-Bench 8K}
& \multicolumn{3}{c}{MME-RW-Lite}
& \multicolumn{3}{c}{MathVista}
& \multicolumn{3}{c}{MathVerse} \\
\cmidrule(lr){2-4}
\cmidrule(lr){5-7}
\cmidrule(lr){8-10}
\cmidrule(lr){11-13}
\cmidrule(lr){14-16}
\cmidrule(lr){17-19}
& $\ell_s$ & $\ell_t$ & $\alpha$
& $\ell_s$ & $\ell_t$ & $\alpha$
& $\ell_s$ & $\ell_t$ & $\alpha$
& $\ell_s$ & $\ell_t$ & $\alpha$
& $\ell_s$ & $\ell_t$ & $\alpha$
& $\ell_s$ & $\ell_t$ & $\alpha$ \\
\midrule
Qwen2.5-VL-7B
& 8 & 8 & 1.0
& 8 & 8 & 1.0
& 9 & 9 & 1.0
& 9 & 9 & 1.0
& 9 & 9 & 1.0
& 9 & 9 & 1.0 \\
Qwen3-VL-4B
& 8 & 14 & 0.5
& 8 & 13 & 0.5
& 9 & 7 & 0.5
& 9 & 8 & 0.5
& 9 & 13 & 0.5
& 9 & 10 & 0.5 \\
InternVL2.5-8B
& 8 & 19 & 0.5
& 8 & 15 & 0.5
& 9 & 20 & 0.5
& 9 & 18 & 0.5
& 9 & 14 & 0.5
& 9 & 18 & 0.5 \\
\bottomrule
\end{tabular}
\end{table*}

\begin{table*}[t!]
\centering
\small
\setlength{\tabcolsep}{6.2pt}
\caption{
    \textbf{Hyperparameter settings of Command-V~\citep{wang2026commandv} .}
    $\ell$ and $\alpha$ denote the injection layer and injection strength, respectively.
}
\label{tab:hyperparameters_commandv}
\begin{tabular}{l*{2}{>{\centering\arraybackslash}p{0.050\textwidth}}*{2}{>{\centering\arraybackslash}p{0.050\textwidth}}*{2}{>{\centering\arraybackslash}p{0.055\textwidth}}*{2}{>{\centering\arraybackslash}p{0.050\textwidth}}*{2}{>{\centering\arraybackslash}p{0.050\textwidth}}}
\toprule
\multirow{2}{*}{Target model}
& \multicolumn{2}{c}{V*}
& \multicolumn{2}{c}{HR-Bench}
& \multicolumn{2}{c}{MME-RW-Lite}
& \multicolumn{2}{c}{MathVista}
& \multicolumn{2}{c}{MathVerse} \\
\cmidrule(lr){2-3}
\cmidrule(lr){4-5}
\cmidrule(lr){6-7}
\cmidrule(lr){8-9}
\cmidrule(lr){10-11}
& $\ell$ & $\alpha$
& $\ell$ & $\alpha$
& $\ell$ & $\alpha$
& $\ell$ & $\alpha$
& $\ell$ & $\alpha$ \\
\midrule
Qwen2.5-VL-7B
& 12 & 2.0
& 12 & 2.0
& 12 & 2.0
& 12 & 2.0
& 12 & 2.0 \\
Qwen3-VL-4B
& 3 & 2.0
& 5 & 2.0
& 5 & 2.0
& 9 & 2.0
& 9 & 2.0 \\
InternVL2.5-8B
& 10 & 2.0
& 10 & 2.0
& 10 & 2.0
& 10 & 2.0
& 10 & 2.0 \\
\bottomrule
\end{tabular}
\end{table*}

\subsection{More details}

\noindent\textbf{More details on capability vector extraction data.}
For capability vector extraction, we sample 1,000 examples from DeepEyes-Datasets-47k~\citep{zheng2026deepeyes}, preserving its source proportions: 475 from the V*~\citep{wu2024datasetsVstar} training split, 290 from ArxivQA~\citep{li-etal-2024-multimodal-arxiv}, and 235 from ThinkLite-VL~\citep{wang2025sota}. In terms of visual content, the subset comprises real-world scenes and objects (543 examples), scientific and mathematical imagery (453 examples), and other visual content (4 examples).

\noindent\textbf{Inference and Evaluation.}
For all evaluations, we use \textbf{greedy decoding}, making the reported results deterministic in principle. We support multi-turn visual interaction and allow at most four interaction rounds for each example. All models are evaluated using the Hugging Face \texttt{Transformers} inference framework with a batch size of 1. To ensure a fair comparison, we keep the inference and evaluation settings identical across all methods and individual experimental runs.
\section{Additional Experiments}
\label{appendix-exp}

\begin{table*}[t!]
\centering
\small
\setlength{\tabcolsep}{2.5pt}
\caption{
    \textbf{Additional results on high-resolution visual understanding benchmarks.}
    We transfer visual-CoT capabilities from DeepEyes-7B to Qwen2.5-VL-3B and Qwen3-VL-32B. We compare against the base target LVLMs (Vanilla), L2V-CoT, Command-V, and Transport and Merge on V* and HR-Bench.
}
\label{tab:additional_main_results_hr}
\begin{tabular}{llccccccccc}
\toprule
\multirow{2}{*}{Model}
& \multirow{2}{*}{Method}
& \multicolumn{3}{c}{V$^*$}
& \multicolumn{3}{c}{HR-Bench 4K}
& \multicolumn{3}{c}{HR-Bench 8K} \\
\cmidrule(lr){3-5}
\cmidrule(lr){6-8}
\cmidrule(lr){9-11}
& & Overall & Attr. & Spatial
& Overall & FSP & FCP
& Overall & FSP & FCP \\
\midrule

DeepEyes-7B
& Base
& 88.48 & 90.44 & 85.53
& 74.00 & 91.25 & 56.75
& 71.13 & 84.50 &57.75 \\
\midrule

\multirow{5}{*}{Qwen2.5-VL-3B}
& Vanilla
& 54.45 & 53.91 & 55.26
& 48.13 & 50.25 & 46.00
& 41.75 & 45.75 & 37.75 \\
& L2V-CoT
& \underline{63.35} & \underline{65.21} & \textbf{60.53}
& 44.50 & 49.50 & 39.50
& \underline{45.63} & \underline{50.00} & \underline{41.25} \\
& Command-V
& 57.59 & 56.52 & 59.21
& \underline{49.13} & \underline{52.00} & \underline{46.25}
& 43.75 & 48.25 & 39.25 \\
& Transport and Merge
& 53.93 & 53.91 & 53.95
& 47.25 & 50.75 & 43.75
& 39.38 & 43.75 & 35.00 \\
\rowcolor{tabhighlight}
\cellcolor{white} & Ours
& \textbf{66.49} & \textbf{70.43} & \textbf{60.53}
& \textbf{58.25} & \textbf{68.25} & \textbf{48.25}
& \textbf{48.75} & \textbf{52.50} & \textbf{45.00} \\
\midrule

\multirow{5}{*}{Qwen3-VL-32B}
& Vanilla
& 92.15 & 92.17 & 92.11
& 86.75 & \textbf{94.25} & 79.25
& 80.88 & 90.00 & 71.75 \\
& L2V-CoT
& \underline{93.72} & \underline{93.91} & \textbf{93.42}
& \underline{87.00} & 93.75 & \underline{80.25}
& 82.88 & 89.50 & 76.25 \\
& Command-V
& \underline{93.72} & \underline{93.91} & \textbf{93.42}
& 86.75 & 93.50 & 80.00
& 82.00 & \underline{90.25} & 73.75 \\
& Transport and Merge
& 91.10 & 93.04 & 88.16
& \textbf{88.00} & 93.75 & \textbf{82.25}
& \textbf{84.00} & 87.50 & \textbf{80.50}\\
\rowcolor{tabhighlight}
\cellcolor{white} & Ours
& \textbf{94.24} & \textbf{94.78} & \textbf{93.42}
& 86.75 & \underline{94.00} & 79.50
& \underline{83.75} & \textbf{90.75} & \underline{76.75} \\
\bottomrule
\end{tabular}
\end{table*}

\begin{table*}[t!]
\centering
\small
\setlength{\tabcolsep}{2.2pt}
\caption{
    \textbf{Additional results on MME-RealWorld-Lite.}
}
\label{tab:additional_main_results_mme}
\begin{tabular}{llcccccccccc}
\toprule
\multirow{2}{*}{Model}
& \multirow{2}{*}{Method}
& \multirow{2}{*}{Overall}
& \multicolumn{5}{c}{Perception}
& \multicolumn{4}{c}{Reasoning} \\
\cmidrule(lr){4-8}
\cmidrule(lr){9-12}
& & & OCR & RS & DT & MO & AD
& OCR & DT & MO & AD \\
\midrule

DeepEyes-7B
& Base
& 52.27
& 90.00 & 50.67 & 89.00 & 44.51 & 34.00
& 73.00 & 70.00 & 46.00 & 35.00 \\
\midrule

\multirow{5}{*}{Qwen2.5-VL-3B}
& Vanilla
& 33.72
& 68.80 & 31.33 & 44.00 & 25.71 & 29.14
& \underline{58.00} & 33.00 & \underline{28.67} & 16.50 \\
& L2V-CoT
& \underline{35.64}
& \underline{69.60} & \textbf{36.00} & \underline{49.00} & 25.39 & \underline{32.29}
& \textbf{59.00} & \underline{38.00} & \textbf{30.00} & \underline{17.75} \\
& Command-V
& 33.98
& 69.20 & \underline{34.00} & 44.00 & 25.71 & 29.14
& \underline{58.00} & 33.00 & \underline{28.67} & 16.50 \\
& Transport and Merge
& 31.74
& 65.60 & 27.33 & 35.00 & \underline{26.33} & 27.43
& 55.00 & 34.00 & 25.33 & 15.50 \\
\rowcolor{tabhighlight}
\cellcolor{white} & Ours
& \textbf{39.71}
& \textbf{70.40} & 24.00 & \textbf{59.00} & \textbf{26.96} & \textbf{37.43}
& 57.00 & \textbf{52.00} & 26.67 & \textbf{31.25} \\
\midrule

\multirow{5}{*}{Qwen3-VL-32B}
& Vanilla
& 56.90
& 90.80 & 58.00 & \underline{91.00} & 41.69 & 43.43
& 83.00 & \textbf{91.00} & \textbf{54.67} & 36.50 \\
& L2V-CoT
& 56.80
& \textbf{92.00} & \textbf{60.00} & \underline{91.00} & 41.38 & 42.00
& \underline{84.00} & 87.00 & 50.67 & \underline{38.25} \\
& Command-V
& 56.70
& 90.80 & 55.33 & \underline{91.00} & 41.69 & 43.43
& 83.00 & \textbf{91.00} & \textbf{54.67} & 36.50 \\
& Transport and Merge
& \underline{57.63}
& \underline{91.20} & \underline{59.33} & \textbf{92.00} & \underline{43.26} & \textbf{45.71}
& 80.00 & 88.00 & 50.00 & \textbf{39.00} \\
\rowcolor{tabhighlight}
\cellcolor{white} & Ours
& \textbf{57.69}
& \underline{91.20} & 57.33 & 88.00 & \textbf{45.14} & \underline{44.57}
& \textbf{86.00} & 90.00 & 52.00 & 37.75 \\
\bottomrule
\end{tabular}
\end{table*}

\begin{table*}[t!]
\centering
\small
\setlength{\tabcolsep}{2.6pt}
\caption{
\textbf{Additional results on mathematical visual reasoning benchmarks.}
}
\label{tab:additional_main_results_math}

\begin{tabular}{llccccccccc}
\toprule
\multirow{2}{*}{Model}
& \multirow{2}{*}{Method}
& \multicolumn{3}{c}{MathVista}
& \multicolumn{6}{c}{MathVerse}
\\
\cmidrule(lr){3-5}
\cmidrule(lr){6-11}
&
& Overall & General & Math
& Overall & TD & TL & VI & VD & VO
\\
\midrule

DeepEyes-7B
& Base
& 66.5 & 69.35 & 64.07
& 43.35 & 50.76 & 44.67 &42.01 & 39.72&39.59
\\
\midrule

\multirow{5}{*}{Qwen2.5-VL-3B}
& Vanilla
& 51.9 & 60.87&44.26
& 32.11 &37.44 &33.12 &30.84 &29.95 &\textbf{29.19}
\\
& L2V-CoT
& \textbf{54.3} &\underline{62.39} &\textbf{47.41}
& 31.93 &\textbf{38.45} &31.85 &\underline{31.09} &30.96 &27.28
\\
& Command-V
& 52.5 & 61.30&45.00
& 32.03 & 36.42& \underline{33.38}& 30.84&30.58 &28.93
\\
& Transport and Merge
& 52.2 & 59.35&\underline{46.11}
& \underline{32.64} &\textbf{38.45} &\underline{33.38} &\underline{31.09} &\underline{31.47} &28.81
\\
\rowcolor{tabhighlight}
\cellcolor{white} &Ours
& \underline{53.9} &\textbf{64.13} &45.19
& \textbf{33.07} & 37.31&\textbf{34.77} &\textbf{31.60} &\textbf{32.49} &\textbf{29.19}
\\
\midrule

\multirow{5}{*}{Qwen3-VL-32B}
& Vanilla
& 78.1 &77.17 &78.89
& \underline{59.57} &\underline{66.24} & \underline{60.79}& \underline{55.84}&55.58 &\underline{59.39}
\\
& L2V-CoT
& \textbf{79.4} &77.39 &\textbf{81.11}
& 58.81 &65.86 &59.90 &54.95 &55.33 &57.99
\\
& Command-V
& \underline{78.4} &\textbf{78.04} &78.70
& 59.26 &\textbf{67.01} & 60.28&54.95 &\underline{55.84} &58.25
\\
& Transport and Merge
& 78.1 &\underline{77.61} &78.52
& 59.04 & 65.23&60.03 &55.58 &55.46 &58.88
\\
\rowcolor{tabhighlight}
\cellcolor{white} &Ours
& 78.1 & 76.52&\underline{80.56}
& \textbf{59.90} &55.84 &\textbf{61.42} &\textbf{56.35} &\textbf{58.63} &\textbf{67.26}
\\
\bottomrule
\end{tabular}
\end{table*}

\noindent\textbf{Additional target models.}
Tables~\ref{tab:additional_main_results_hr}, \ref{tab:additional_main_results_mme}, and \ref{tab:additional_main_results_math} report results for Qwen2.5-VL-3B and Qwen3-VL-32B.
In particular, transferring capabilities from DeepEyes-7B improves Qwen3-VL-32B's V* accuracy from 92.15\% to 94.24\%, showing that \ours~remains effective for substantially larger target LVLMs.

\noindent\textbf{Additional results on capability disentanglement.}
Table~\ref{tab:additional_ablation_gr} reports additional capability disentanglement results on InternVL2.5-8B. Combining the grounding and reasoning vectors achieves the best overall performance on V*, MME-RealWorld-Lite, and MathVista, outperforming the unified visual-CoT vector on all three benchmarks. This result further validates the effectiveness of capability disentanglement across different target models.

\noindent\textbf{Additional results on phase-aware routing.}
Table~\ref{tab:additional_ablation_routing} reports additional phase-aware routing results on InternVL2.5-8B. Removing routing reduces V* accuracy from 72.25\% to 62.30\%, below the vanilla model's 69.63\%. This result further validates the effectiveness of phase-aware routing across different target models.

\begin{table}[t!]
\centering
\small
\setlength{\tabcolsep}{6pt}
\caption{
    \textbf{Additional ablation study of disentangled capability vectors on InternVL2.5-8B.}
    $\widetilde{\mathbf{v}}_g$ and $\widetilde{\mathbf{v}}_r$ denote the grounding and reasoning capability vectors, respectively, while $\widetilde{\mathbf{v}}$ denotes the unified visual-CoT capability vector.
}
\label{tab:additional_ablation_gr}

\begin{tabular}{lccccccccc}
\toprule
\multirow{2}{*}{Method}
& \multicolumn{3}{c}{V*}
& \multicolumn{3}{c}{MME-RealWorld-Lite}
& \multicolumn{3}{c}{MathVista}
\\
\cmidrule(lr){2-4}
\cmidrule(lr){5-7}
\cmidrule(lr){8-10}
& Overall & Attr. & Spatial
& Overall & Percep. & Reason.
& Overall & General & Math
\\
\midrule

Vanilla
& 69.63 & 66.96 & 73.68
& 44.09 & 47.48 & 38.80
& 58.0 & 61.74 & 54.81
\\

\quad$+\widetilde{\mathbf{v}}_g$
& \underline{70.68} & \underline{67.83} & \underline{75.00}
& 44.14 & 46.96 & 39.73
& 58.3 & 60.87 & 56.11
\\

\quad$+\widetilde{\mathbf{v}}_r$
& \underline{70.68} & \underline{67.83} & \underline{75.00}
& 44.92 & \underline{48.33} & 39.60
& 59.9 & 64.78 & 55.74
\\

\quad$+\widetilde{\mathbf{v}}$
& 70.16 & \underline{67.83} & 73.68
& \underline{45.23} & 48.16 & \underline{40.67}
& \underline{60.0} & 63.48 & 57.04
\\

\rowcolor{tabhighlight}
\quad$+\widetilde{\mathbf{v}}_g, +\widetilde{\mathbf{v}}_r$
& \textbf{72.25} & \textbf{69.57} & \textbf{76.32}
& \textbf{46.17} & \textbf{49.27} & \textbf{41.33}
& \textbf{61.5} & \textbf{65.22}  & \textbf{58.32}
\\

\bottomrule
\end{tabular}
\end{table}

\makeatletter
\begin{table*}[t!]
\noindent
\begin{minipage}[t]{0.48\textwidth}
\vspace{0pt}
\centering
\def\@captype{table}
\small
\setlength{\tabcolsep}{8pt}
\caption{
    \textbf{Additional results on phase-aware routing with InternVL2.5-8B.}
}
\label{tab:additional_ablation_routing}
\begin{tabular}{lccc}
\toprule
\multirow{2}{*}{Method}
& \multicolumn{3}{c}{V*}
\\
\cmidrule(lr){2-4}
& Overall & Attr. & Spatial
\\
\midrule
Vanilla
& \underline{69.63} & \underline{66.96} & \underline{73.68}
\\
w/o Routing
& 62.30 & 61.74 & 63.16
\\
\rowcolor{tabhighlight}
Ours
& \textbf{72.25} & \textbf{69.57} & \textbf{76.32}
\\
\bottomrule
\end{tabular}
\vspace{-7pt}
\setlength{\tabcolsep}{6.3pt}
\caption{
    \textbf{Alternative alignment.}
}
\label{tab:projection_alignment}
\begin{tabular}{lccc}
\toprule
\multirow{2}{*}{Method}
& \multicolumn{3}{c}{V*}
\\
\cmidrule(lr){2-4}
& Overall & Attr. & Spatial
\\
\midrule
Vanilla & 86.39 & 89.57 & 81.58 \\
LMN & \textbf{92.67} & \textbf{93.91} & \textbf{90.79} \\
Projection-based
& \textbf{92.67} & \textbf{93.91} & \textbf{90.79} \\
\bottomrule
\end{tabular}
\end{minipage}\hfill
\begin{minipage}[t]{0.48\textwidth}
\centering
\def\@captype{table}
\small
\setlength{\tabcolsep}{5.3pt}
\caption{
    \textbf{Effect of \ours~on bounding-box IoU and response length on V* using Qwen2.5-VL-7B.}
    Resp. Len. is the mean number of tokens in the model-generated trajectory; IoU denotes the mean IoU of the bounding box.
    $\Delta$ denotes the change from Vanilla to \ours.
}
\vspace{6pt}
\label{tab:bbox_length}
\begin{tabular}{lc|cc}
\toprule
Method & Acc. & Resp. Len. & IoU \\
\midrule
DeepEyes-7B & 88.48 & 835.8 & 0.3911 \\
\midrule
Vanilla & 75.92 & 98.9 & 0.3102 \\
\quad$+\widetilde{\mathbf{v}}_g$ & 86.39 & 100.8 & 0.3445 \\
\quad$+\widetilde{\mathbf{v}}_r$ & 86.39 & 161.1 & 0.2984 \\
\rowcolor{tabhighlight}
\quad$+\widetilde{\mathbf{v}}_g, +\widetilde{\mathbf{v}}_r$ & \textbf{93.72} & \textbf{162.0} & \textbf{0.3472} \\
\rowcolor{tabhighlight}
\quad$\Delta$ & +17.80 & +63.1 & +0.0370 \\
\bottomrule
\end{tabular}
\end{minipage}
\end{table*}
\makeatother

\noindent\textbf{Bounding-box overlap and response length.}
Table~\ref{tab:bbox_length} compares grounding quality and response length on V* using Qwen2.5-VL-7B.
Grounding steering improves grounding precision, while reasoning steering produces deeper and more accurate reasoning traces. Combining both vectors yields an IoU of 0.3472 with an average response length of 162.0 tokens. These results demonstrate that \ours~jointly enhances precise visual grounding and high-quality reasoning.

\noindent\textbf{Alternative alignment.} To verify that our gains do not depend on the frequency-domain LMN component, we replace it with a parameter-free \emph{Projection} alignment while keeping all other settings fixed.  For each capability $c\in\{g,r\}$, let $\mathbf{X}_s^c\in\mathbb{R}^{N\times d_s}$ and $\mathbf{X}_t^c\in\mathbb{R}^{N\times d_t}$ be the centered, paired activation-difference matrices, and let $\mathbf{V}_s^c$ and $\mathbf{V}_t^c$ be their complete sample-supported PCA bases with $p=\min(N-1,d_s,d_t)$ components.  Projection aligns the paired PCA coordinates with an orthogonal map~\citep{Schönemann_1966} and transports the source vector as
\begin{equation}
    \mathbf{R}_c^*
    =\underset{\mathbf{R}^{\top}\mathbf{R}=\mathbf{I}}{\arg\min}
    \left\lVert\mathbf{X}_s^c\mathbf{V}_s^c\mathbf{R}-\mathbf{X}_t^c\mathbf{V}_t^c\right\rVert_F^2,
    \qquad
    \widetilde{\mathbf{v}}_c
    =\mathbf{V}_t^c(\mathbf{R}_c^*)^{\top}(\mathbf{V}_s^c)^{\top}\mathbf{v}_c.
    \label{eq:projection_alignment}
\end{equation}
As shown in Table~\ref{tab:projection_alignment}, Projection-based alignment achieves the same performance as LMN on V* using Qwen3-VL-4B under identical experimental settings, confirming that the effectiveness of our capability-transfer framework is not tied to a particular cross-architecture alignment operator.

\begin{wraptable}{r}{0.42\columnwidth}
\centering
\small
\setlength{\tabcolsep}{4.8pt}
\vspace{-15pt}
\caption{\textbf{Inference latency on V* using Qwen3-VL-4B.}}
\label{tab:inference_latency}
\begin{tabular}{@{}lccc@{}}
\toprule
Method & Acc. & ms/token & Latency ratio \\
\midrule
Vanilla & 86.39 & 19.96 & $1.00\times$ \\
\rowcolor{tabhighlight}
Ours & \textbf{92.67} & 21.77 & $1.09\times$ \\
\bottomrule
\end{tabular}
\end{wraptable}
\noindent\textbf{Inference latency.}
We measure the inference latency of \ours~and the vanilla model on V* using Qwen3-VL-4B and a single NVIDIA H100 80GB GPU with a batch size of 1. Under otherwise identical settings, the vanilla model requires 19.96 ms per generated token, whereas \ours~requires 21.77 ms per token, corresponding to a latency ratio of only $1.09\times$. This result shows that \ours~introduces negligible inference overhead while substantially improving accuracy.

\end{document}